\documentclass[journal]{IEEEtran}

\usepackage{amsmath}
\usepackage{xcolor}
\usepackage{bbold}
\usepackage{graphicx}
\usepackage{makecell}
\usepackage{multirow}
\usepackage{booktabs}
\usepackage{tabu}
\usepackage{float}
\usepackage{subcaption}

\usepackage[ruled,vlined]{algorithm2e}
\newcommand{\argmin}{\arg\!\min}

\DeclareCaptionLabelFormat{andtable}{#1~#2  \&  \tablename~\thetable}

\begin{document}

\title{A Deep Learning Approach for Macroscopic Energy Consumption Prediction with Microscopic Quality for Electric Vehicles}

\author{Ayman~Moawad,
        Krishna~Murthy~Gurumurthy,
        Omer~Verbas,
        Zhijian~Li,
        Ehsan~Islam,
        Vincent~Freyermuth,
        and~Aymeric~Rousseau% <-this % stops a space
\thanks{A. Moawad is with the Vehicle and Mobility Simulation group at Argonne National Laboratory, 9700 S Cass Ave, Lemont, IL 60439 USA and the department of Statistics, The University of Chicago, 5801 S Ellis Ave, Chicago, IL 60637 USA. E-mail: amoawad@anl.gov, aymoawad@uchicago.edu}% <-this % stops a space
\thanks{Krishna Murthy Gurumurthy, Omer Verbas, Ehsan Islam, Vincent Freyermuth, and Aymeric Rousseau, are with the Vehicle and Mobility Simulation group at Argonne National Laboratory.}% <-this % stops a space
\thanks{Zhijian Li is with the department of Mathematics at The University of California Irvine.}% <-this % stops a space
}

\maketitle

\begin{abstract}
This paper presents a machine learning approach to model the electric consumption of electric vehicles at macroscopic level, i.e., in the absence of a speed profile, while preserving microscopic level accuracy. For this work, we leveraged a high-performance, agent-based transportation tool to model trips that occur in the Greater Chicago region under various scenario changes, along with physics-based modeling and simulation tools to provide high-fidelity energy consumption values. The generated results constitute a \textit{very} large dataset of vehicle-route energy outcomes that capture variability in vehicle and routing setting, and in which high-fidelity time series of vehicle speed dynamics is masked. We show that although all internal dynamics that affect energy consumption are masked, it is possible to learn aggregate-level energy consumption values quite accurately with a deep learning approach. When large-scale data is available, and with carefully tailored feature engineering, a well-designed model can overcome and retrieve latent information. This model has been deployed and integrated within POLARIS\footnote{POLARIS is an Argonne-based high-performance, open-source agent-based modeling framework designed to simulate large-scale transportation systems.} Transportation System Simulation Tool to support real-time behavioral transportation models for individual charging decision-making, and rerouting of electric vehicles.

\end{abstract}


\IEEEpeerreviewmaketitle

\section{Introduction}
\IEEEPARstart{T}{he} energy consumption impacts of battery electric vehicles (EVs) have been well studied under standard drive cycles (\cite{moawad_assessment_2016}, \cite{islam_extensive_2018}, \cite{islam_energy_2020}, \cite{islam_energy_2021}), and have shown promise as a solution to reduce emissions and dependence on petroleum. Although regulatory cycles are a good way to standardize energy benefit studies, they fail to represent real driving conditions and traffic variability. Therefore, recent research focuses on the predictability of energy consumption in EVs in more realistic settings, which would yield large impacts in many application fields, from driver range-anxiety studies, routing choices, eco-routing, and fleet management of shared vehicles or freight, to power grid and charging infrastructure requirements.

\begin{figure}[ht!]
\centering
\includegraphics[width=3.4in, clip]{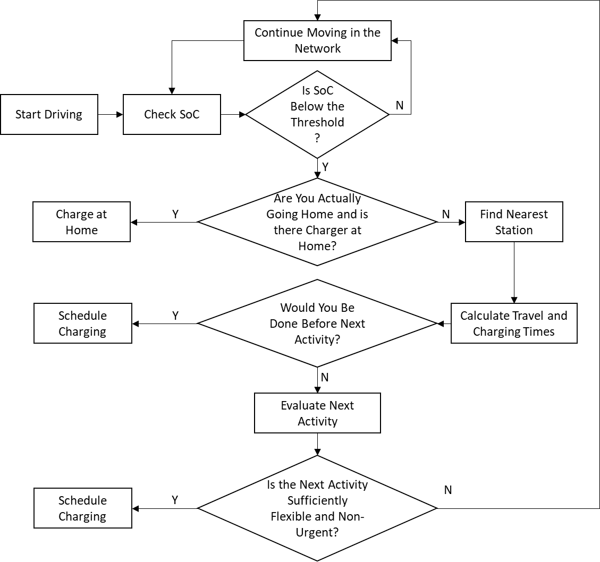}
\caption{Example flowchart of charging behavior scheduling/planning by transportation system agents. In this flow, an embedded real-time energy consumption prediction model helps anticipate SOC depletion in upcoming links of a trip. Charging decisions can be planned accordingly for every agent.
}
\label{fig:CB}
\end{figure}

The driving and charging behavior of EV drivers is key to properly modeling an entire transportation system. Because EV penetration is expected to increase throughout the system, charging loads, needs, and routing decisions will change significantly. For example, the charging behavior of medium- and heavy-duty trucks (MD/HDTs) and transportation network company (TNC) vehicles can significantly disrupt the network, affecting EV charging station capacity and location. This new trend calls for better planning, namely by estimating EV charging demand needs under heterogeneous driving behaviors, various traffic conditions, and a variety of scenarios and EV technologies to tackle the true scale of the impact. A series of tools developed at Argonne help provide a pathway to simulate these high quality data. In section \ref{sec:DG}, we provide some details about our experimental design and the data generation process to address such a diversified setup.
One critical component in identifying charging demand is tracking the state of charge (SOC) of a vehicle's battery (Figure \ref{fig:CB}). Battery SOC evolution over a given trip, and generally throughout the day, is highly influenced by the battery capacity, initial SOC (i.e., the initial energy content of the battery), the anticipated route, and the subsequent dynamics of the vehicle. Therefore, in this work we focus on the predictability of battery energy depletion by looking at link-level (road segment) consumption over an expected trip. By accurately understanding battery consumption over large regions, transportation demand modeling tools can identify charging needs and predict when travelers are likely to schedule their charging trip based on their daily set of activities. We leverage the agent-based transportation tool POLARIS (\cite{auld2016polaris}) to model all trips, activities, and vehicle distribution across the Greater Chicago region under various scenario changes. This data is then paired with the physics-based modeling and simulation tool Autonomie\footnote{Autonomie is a MATLAB-based software environment and framework for automotive modeling, control system design, simulation, and analysis.} to produce large-scale energy consumption estimates. The results we generate serve as a very large backbone dataset of vehicle, route and energy outcomes that capture variability in vehicle classes, powertrain fleet distribution, vehicle technology, automation and connectivity levels, population, driving modes, ride sharing magnitude, e-commerce impact, and other factors, all of which impact traffic and driving behavior.

In section II and III, the development of a machine learning approach to efficiently estimate the link-level energy consumption of EVs over various routes using this dataset is described. We define a route as a sequence of links in which little internal dynamics information is known (i.e., vehicle speed dynamics is masked). In section IV, we summarize model results and argue that, when large-scale, rich, diversified, and carefully designed trip data is available, it is possible to learn aggregate-level energy consumption values quite accurately without high-fidelity speed profiles. As shown in Figure \ref{fig:CB}, with deployment goals in mind, the model needs to be lightweight, efficient, and scalable for subsequent integration into transportation systems, allowing real-time (on-the-road) querying of energy consumption and anticipated SOC, leading to a better understanding of charging demand across any given region. In section V, the deployment of the model developed is shown for the case of POLARIS with preliminary metrics of prediction accuracy and computational load with and without querying hundreds of thousands of predictions.

\section{Data and Learning Setup}
\label{sec:DLS}

\subsection{Data Generation Process}
\label{sec:DG}
\begin{figure}[ht!]
\centering
\includegraphics[width=\linewidth]{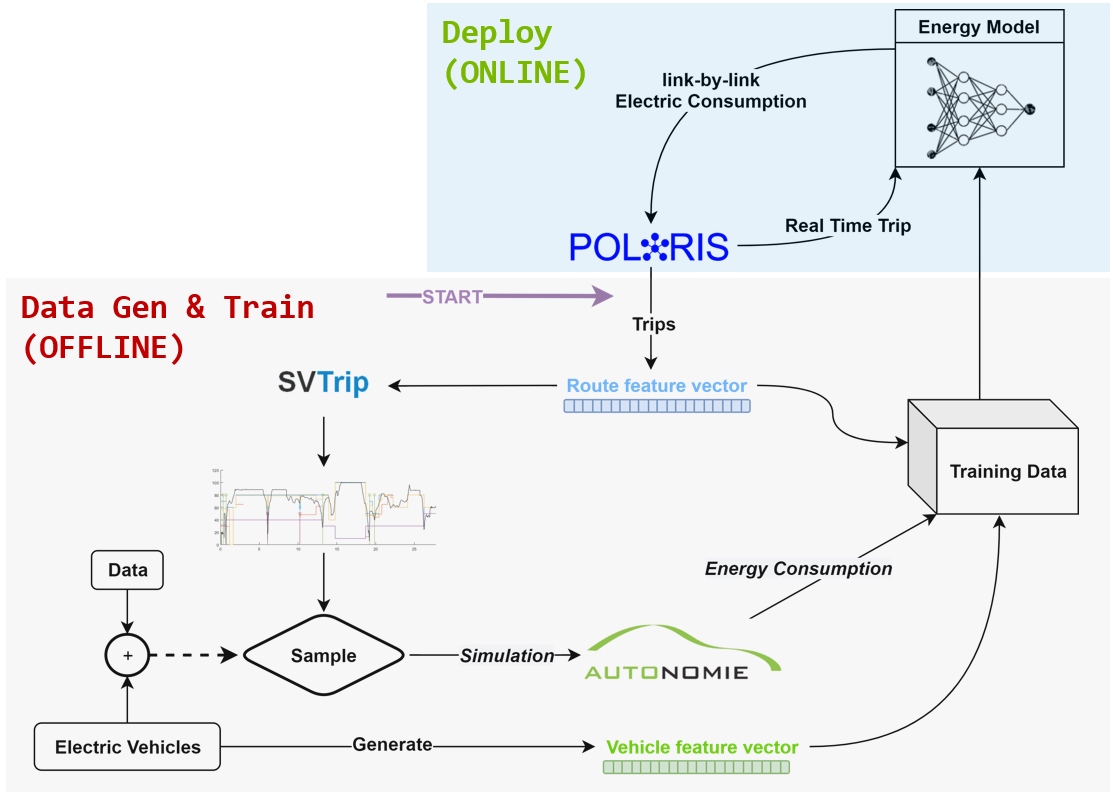}
\caption{Data Generation and Training Process (grey box). After training, the model is deployed as a standalone model and integrated into POLARIS Transportation System to make on-demand queries of energy needs, and anticipate SOC status for new simulated agents (light blue box).}
\label{fig:dgp}
\end{figure}

In order to determine the energy consumption of a metropolitan area under varying conditions, it is necessary to model the entire transportation system. Activity demand is generated to become the basis for the generation of trips. Trips generate traffic flow, which in turn affects average trip speed. Trip information at the link level can then be used along with powertrain information to inform fuel consumption. Autonomie\footnote{Autonomie is a MATLAB-based software environment and framework for automotive modeling, control system design, simulation, and analysis.} (\cite{freyermuth_energy_2019} and \cite{freyermuth_powertrain_2020}), POLARIS (\cite{auld2016polaris} and \cite{gurumurthy2020integrating}) and SVTrip (\cite{karbowski_trip_2014}) presents a complete workflow to study this complex and multidimensional system, as shown in Figure \ref{fig:dgp}.

POLARIS is used to develop and validate a transportation system model for the Chicago Metropolitan Area. It uses population and vehicle synthesis, calibrated with Census data for the region, along with a complete activity demand generation and scheduling, and subsequent traffic simulation, to model the entire transportation system. On simulating trips in a region, link-level trajectories for all vehicle trips tied with vehicle characteristics is obtained. This route information is fed into SVTrip, which predicts 1-Hz speed profiles for each trip to extract microscopic changes in vehicle movement across a link. The resulting detailed trajectory data is then simulated with Autonomie to estimate the energy consumption of the transportation network for different vehicle technologies. This workflow provides energy estimates in an offline simulation based, high-performance computing setting, \textit{and} in a one-way (POLARIS $\rightarrow$ SVTrip $\rightarrow$ Autonomie), slow and computationally demanding environment. Because of the high fidelity of the different pieces involved in the workflow, this setup provides accurate energy outcomes. However, it is inadequate for online, on-demand energy calls in which the mobility context is constantly changing, and in which real-time decisions need to be made, such as on-road EV charging decisions and rerouting.

\subsection{Masking and Latent Energy Learning}
\begin{figure}[ht!]
\centering
\includegraphics[width=\linewidth]{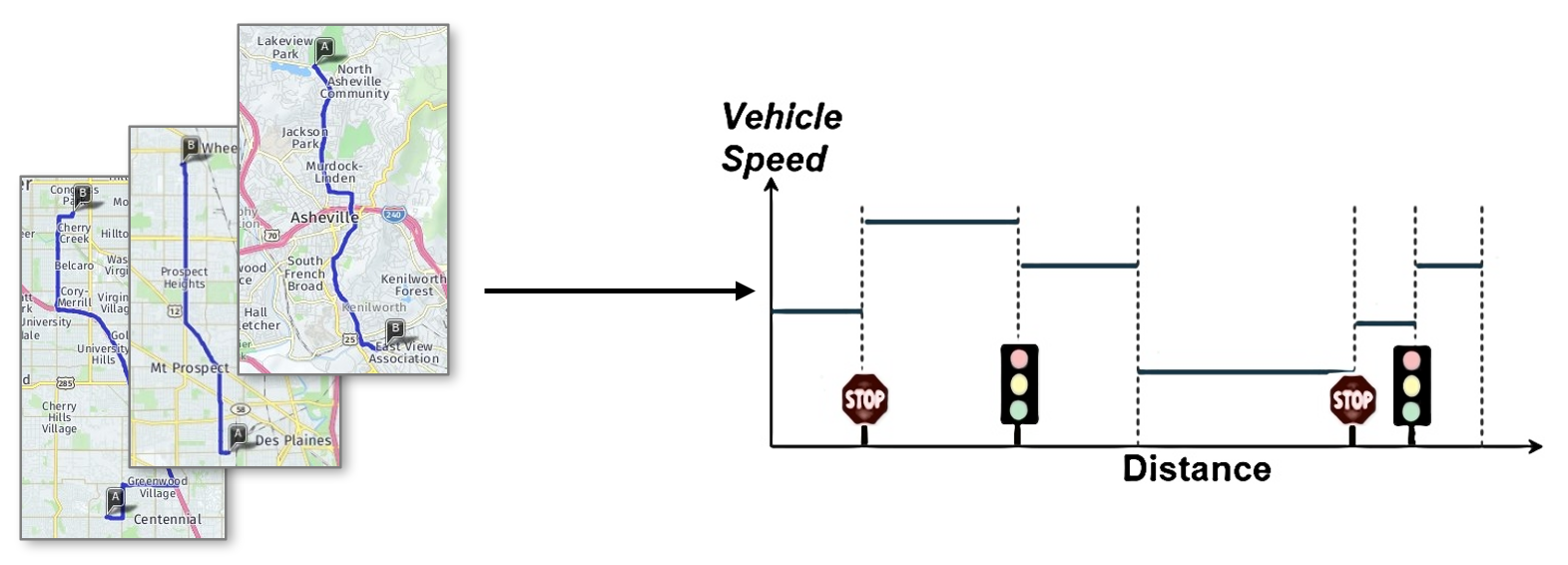}
\caption{Macroscopic level trip information.}
\label{fig:macro_spd}
\end{figure}

The macroscopic-level trips generated from POLARIS in the workflow in Section \ref{sec:DG} contain very basic information (expected average speed over link sequence) about the inner traffic dynamics (Figure \ref{fig:macro_spd}). The goal is to learn energy consumption on that incomplete summary-level information of the dynamics. Typically, high-resolution speed behavior is key to estimating the energy profile. However, in our setting the learning occurs without it: learn the relationship between high-level route structure and vehicle feature interactions to understand how they correlate with the energy consumed. As a result, in the process described in Figure \ref{fig:dgp}, the high-fidelity time series of vehicle speed dynamics is masked. The data used to train the energy model contain vehicle parameters and high-level aggregated route information only (i.e., macroscopic-level data). This approach is consistent with the information available in real-world situations (e.g., route information from HERE or Google maps). Correspondingly, POLARIS provides trajectory information at the link-level. Generated trips include details about the expected link average speed, link length, speed limits, and similar level parameters. Major vehicle attributes are generally considered available and known; if not, they can generally be retrieved as shown in \cite{XAI_moawad} (vehicle weight, battery size, etc.). Therefore, vehicle information is needed and considered as input to the modeling.
Although all internal dynamics that affect energy consumption are masked, we show that it is possible to learn aggregate-level energy consumption values quite accurately with a deep-learning approach. When large-scale data is available, and with some tailored feature engineering, such a model can overcome latent information.

\subsection{The Data}
\label{sec:data}
\begin{figure}[ht!]
\centering
\includegraphics[width=\linewidth]{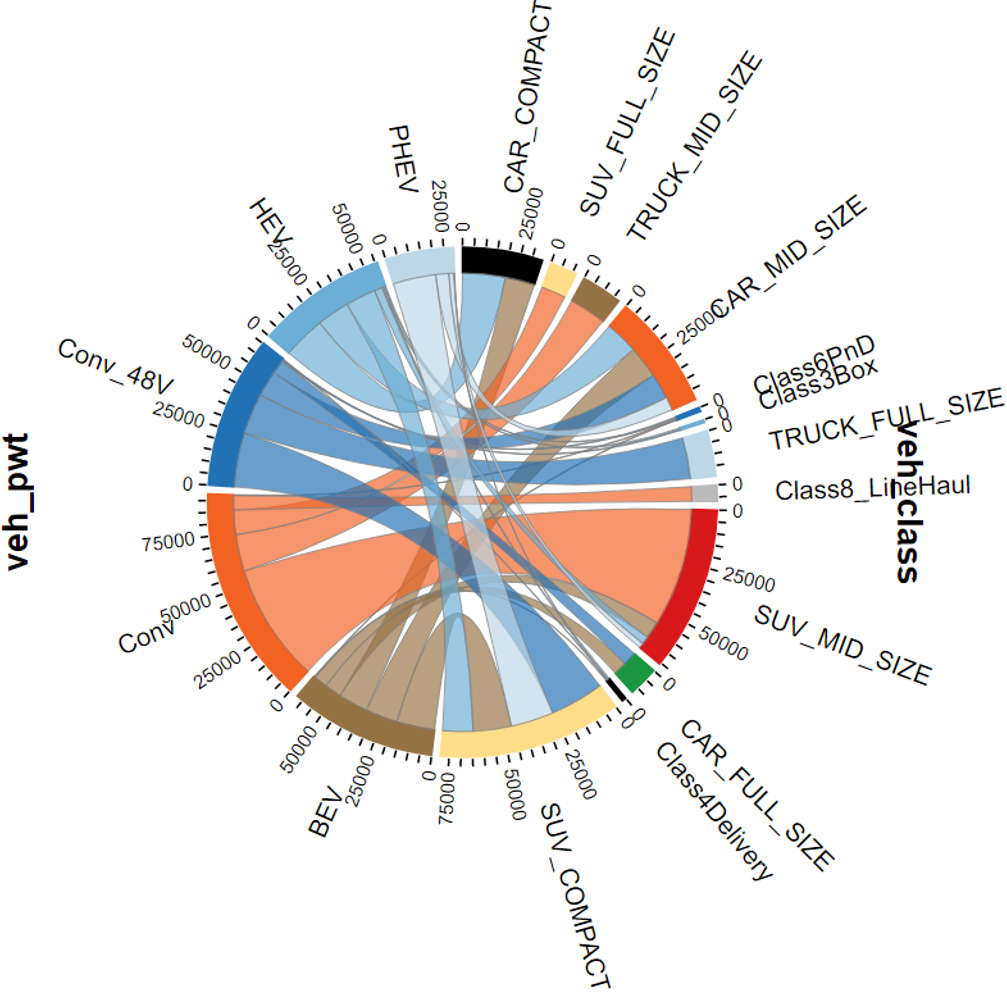}
\caption{Sample fleet distribution count over vehicle powertrains and classes modeled in a scenario. The battery EV data constitutes the core of the training set.}
\label{fig:chord}
\end{figure}

Several scenarios have been defined to cover current and potential future vehicle specifications and transportation behaviors. These scenarios vary in time frame, connectivity and automation, ride sharing, freight systems, and other factors. One scenario reflects today’s situation, where most vehicles are privately owned. Other scenarios represent a short-term future where technology enables people to significantly increase the use of transit, car and ride sharing, and multi-modal travel. Partial automation is introduced, but it is mainly limited to the highway system. Later scenarios represent the long-term future where technology has taken over, enabling high usage of ride sharing and multi-modal trips, which are are convenient and affordable. As a result, private ownership has decreased and e-commerce is common, as is telecommuting. Partially and fully automated vehicles are widely accessible. This extends to scenarios where fully automated vehicles within households are common, and personal ownership results in a low ride-sharing market. The ability to own autonomous vehicles may lead to lower e-commerce and alternative work schedules, and feed urban sprawl.
In each scenario the transportation system as a whole is affected, which causes rich and diverse mobility behavior. In addition, with moving time frames, vehicle technologies develop and diversify; in other words, vehicle specifications, power and efficiency maps, component sizes, and other vehicle aspects are affected.
This design of simulation creates a dataset that covers a large portion of the sample space. It generates enough examples to model the joint distribution over the vehicle and route features, which allows the subsequent deep-learning model to learn systematic relationships and energy outcomes under a variety of cases.
The generated data contains hundred of thousands of EV trips over 30,000+ Chicago links, among more than 3.5 million trips of all types of powertrains (conventional, hybrid vehicles, plug-in hybrids, fuel cells, etc.). It accommodates a wide range of light-duty vehicles, such as compact, midsize, sports utility, pickups, as well as MD/HDT, such as class 3, class 4, class 6, and 8 trucks with different vocations such as pickup and delivery trucks, walk-in, van, box, long-haul, and others. Figure \ref{fig:chord} shows a sample count of powertrains and class distribution. As explained before, battery EVs are modeled with varying electric ranges, component sizes, specifications, automation levels (no automation, partial automation, full automation), and other parameters.

\begin{figure}[!ht]
    \centering
    \includegraphics[width=\linewidth]{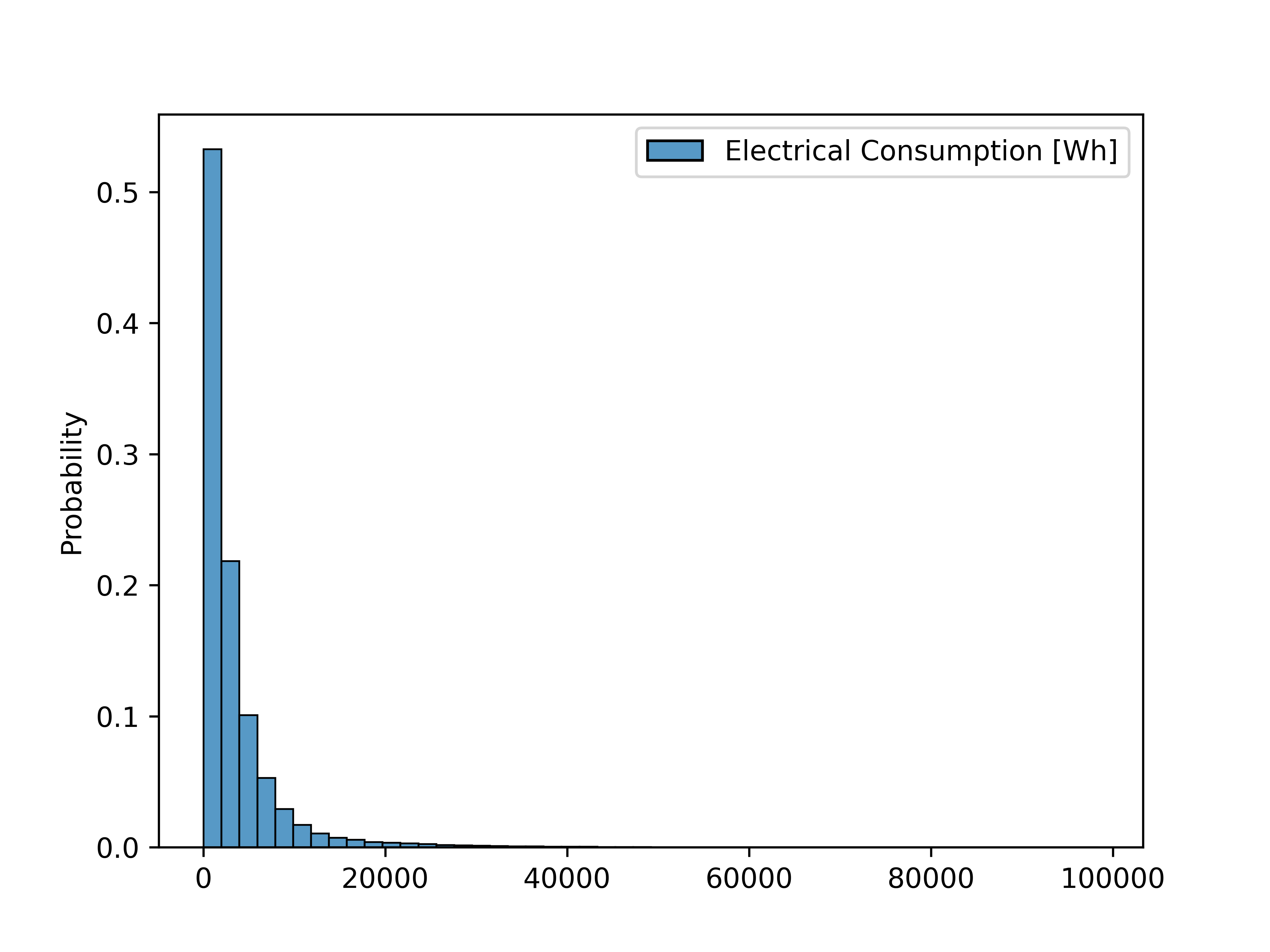}
    \qquad
\resizebox{0.55\linewidth}{!}{%
\begin{tabular}{cc}
                                    & \textbf{Electric Consumption {[}Wh{]}} \\ \hline
\multicolumn{1}{c|}{\textbf{Statistic}} & Value                                 \\ \hline
\multicolumn{1}{c|}{\textbf{Count}} & 342920                                 \\ 
\multicolumn{1}{c|}{\textbf{Mean}}  & {\color[HTML]{000000} 3379.8}          \\
\multicolumn{1}{c|}{\textbf{Std. Deviation}}   & 5084.9                                 \\
\multicolumn{1}{c|}{\textbf{Minimum}}   & 12.6                                   \\
\multicolumn{1}{c|}{\textbf{25\%ile}}  & 695                                    \\
\multicolumn{1}{c|}{\textbf{50\%ile}}  & 1785.9                                 \\
\multicolumn{1}{c|}{\textbf{75\%ile}}  & 3930.6                                 \\
\textbf{Maximum}                        & 98398.3                               
\label{tab:ecstats}
\end{tabular}
}
    \captionsetup{labelformat=andtable}
    \caption{Training data summary statistics. Close to 350,000 EV trips (0.8 split) are sampled from the set of simulated trips.}
    \label{fig:ecdist}
  \end{figure}

Figure \ref{fig:ecdist} shows the distribution and summary statistics of the electric consumption of battery EVs over trips. The electric consumption values are in Wh. We see that the mean is significantly higher than the median, which indicates a large variance of consumption values.

\subsection{Initial Feature Engineering}
\label{sec:FeatEng}
\begin{figure}[ht!]
\centering
\includegraphics[width=\linewidth]{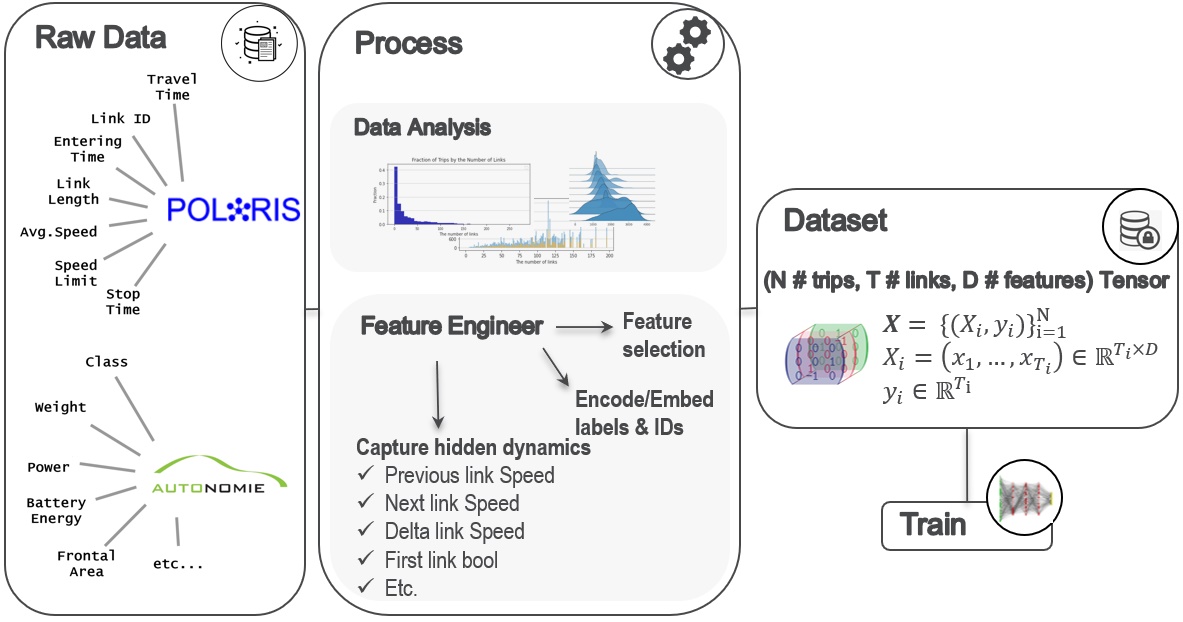}
\caption{Data Construct from raw Autonomie and POLARIS outputs for $N$ trips, sequenced over a series of $T$ links, each described by $D=D_1+D_2$ vehicle and route features. Data is carefully analyzed, processed, and augmented with additional engineered features.}
\label{fig:dataConstruct}
\end{figure}

Let $\mathcal{X}= \{(\mathbf{X}_i,\mathbf{y}_i)\}_{i=1}^N$ be the sample space of data for $N$ trips. $\mathbf{X}_i\in \mathcal{R}^{T_i\times D}$ is a trip $i$ with $T_i$ links and $D$ dimensions so that $\mathbf{X}_i=(\mathbf{x}_1,\cdots, \mathbf{x}_{T_i})$, where $\mathbf{x}_t=[\mathbf{u}_t,\mathbf{v}_t]\in \mathcal{R}^D$ is a link. We define $v_t\in \mathcal{R}^{D_1}$ and $u_t\in \mathcal{R}^{D_2}$ as the vehicle features and link features respectively with $D_1+D_2=D$ dimensional features. The labels $\mathbf{y}_i\in \mathcal{R}^{T_i}$ represent the electric consumption of vehicles on each link. As a result, the data is constructed as a rank 3 tensor of ($N$ trips, $T$ links, $D$ features). The $D$ features are concatenated from the vectors of vehicle and route features, along with manually feature-engineered ones. As explained before, raw route features data from a macroscopic framework, such as in POLARIS, provide very limited route information. Figure \ref{fig:dataConstruct} shows the seven main POLARIS link-level activity outputs. These include a link ID, the link entering time for the vehicle, the link length, the expected stop duration on link, the expected link travel time, the expected average speed on link, and the speed limit on link.
Because vehicles do not enter or exit the same link at the same time or in the same way, similar (or even identical) links do not necessarily yield similar energy profiles. In a given trip, a route is composed of a sequence of links and the energy consumed will depend on what happened in previous links and what is expected to happen on the next links. In other words, the structure of the current link (the focus) along with the surrounding behavior (the context) will dictate how much energy is expected to be consumed. This a priori known fact embodies our intuition to construct and engineer explicit features to recover a portion of such hidden dynamics (e.g., delta speed of surrounding links). Such macroscopic-level dynamics are the aftereffects of more granular microscopic changes. Designed features that capture sequence dependencies will explicitly guide and inform the model training to learn microscopic latent behavior.

In the dataset used, we have $D_1=20$ the dimensionality of the vehicle features, and $D_2=26$ the dimensionality of the link features, so that $D=46$. $D_1$ includes vehicle data such as vehicle class, weight, battery size, frontal area, etc. $D_2$ includes raw route features from macroscopic-level data, in addition to feature engineered ones, such as delta link speeds, delta length, naive congestion calculations comparing average expected speed to speed limits, and other calculated variables.\\

\subsubsection{The Problem of Varying Trip Length}
Trips have different lengths based on traveler, activity, and their home locations. It is common to use zero padding to equalize data points to the same shape. However, in this case, the number of links can vary from 1 to 284, and padding jeopardizes the accuracy of the model by introducing irrelevant data. To better train the model, we experimented with different ways of batching the link sequence data, which had high variance in trip length. In one trial, we grouped the data by similar trip length (with shuffling at the end of each epoch to introduce noise) to minimize padding within a batch. On grouping by similar trip length, we further experimented with quantile bucketing and binning of different sizes. That is, if $T=\max_{1\leq i \leq N}T_i$, we discretize the range of trip length into levels $0= L_0<L_1<\cdots < L_m = T$. Based on this discretization, we partition the dataset into disjoint groups $\mathcal{X}=\bigsqcup_{k=1}^{m}\mathcal{G}_k$, where $X_i\in \mathcal{G}_k$ if $T_i\in (L_{k-1},L_{k}]$. Then, for each group $\mathcal{G}_k$, we map all data points in $\mathcal{G}_k$ into $\mathcal{R}^{L_k\times D}$ through padding zeros.
In another trial, at the expense of less noise during training, we grouped the sequence data by \textit{exactly} the same trip length to completely avoid padding. The latter approach yielded the best result.

\section{The Model}
\subsection{Setup}
Our goal is to build a model that can accurately estimate the electric consumption of EVs under large vehicle, roads, trips, and driving variability. Although we have features that seem to directly impact energy consumption, such as trip length, average speed, vehicle weight, there are significant difficulties in modeling the energy consumption explicitly at the link level. It is hard to capture the interactions between links. For example, since the fuel consumption rates for acceleration and slowing down are very different at the same average speed, a vehicle's energy consumption on a link varies depending on how it enters and leaves the link. A neural network can learn such latent information from a sufficiently large and rich dataset.

\subsection{Overall Structure}
In section \ref{sec:FeatEng}, we manually and explicitly engineered features based on our knowledge of the problem. We design in this section a deep learning model to implicitly account for link dependencies through recurrent neural network (RNN) sequence modeling. In addition, we pre-process the sequence input with an automated feature extraction mechanism leveraging a convolutional neural network (CNN) (\cite{lecun_deep_2015}, \cite{NIPS2012_c399862d}, \cite{SimonyanZ14a}, \cite{7298594}).

\begin{figure*}[ht!]
\includegraphics[width=\textwidth]{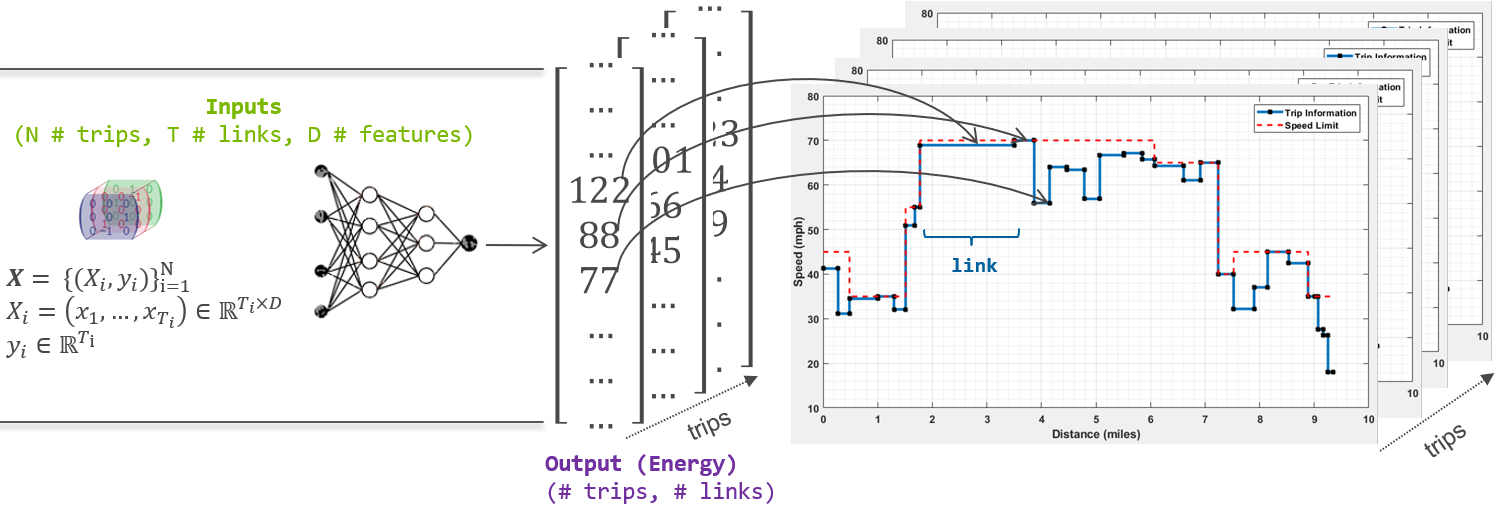}
\caption{Model Inference. The model takes a sequence of links over the trip with concatenated vehicle and route level features, then predicts link-level energy outcomes for a link by estimating link energy and battery SOC. On model training, $\mathbf{X}$ represents a batched and bucketed dataset of trips of same length, $X_i$ provides input data for a given trip, and $y_i$ is the labeled energy outcome (in Wh) from training labeled data. Loss is minimized over the trip, link, and cumulative error.}
\label{fig:model}
\end{figure*}

More specifically, the fuel consumption of a vehicle on a link depends on the preceding and following links. For that modeling portion we leverage a bi-directional RNN with long short-term memory (LSTM) cells \cite{hochreiter1997long} stacked with a fully connected neural network as a series of time-distributed dense layers of varying dimensions. The time-distributed property preserves the output sequence for link-level evaluation of electric consumption (figure \ref{fig:model}). The bi-directional aspect (BiLSTM, \cite{650093}) captures the contextual qualities surrounding the current link. BiLSTMs understand context better because the sequence runs forward and backward, and two internal states summarizing the entire trip are combined. Overall, this portion of the network attempts to learn the sequential dependencies of links and their consumption. A high-level overview of the model is that, given an input batch of trips $\mathbf{X}$ with length $\Tilde{T}$, the output is formulated as 
\begin{equation}
\begin{aligned}
&\mathbf{\Tilde{X}} = CNN(\mathbf{X})\\
&\mathbf{H} = biLSTM(\mathbf{\Tilde{X}})\\
&\Tilde{y}_{rnn}=g(\mathbf{H}w_{rnn}+b_{rnn})
\end{aligned}  
\end{equation}
where for an $h$ the size of the hidden units of the BiLSTM stack, $\mathbf{H}\in \mathcal{R}^{\Tilde{T}\times h}$, $w_{rnn}\in \mathcal{R}^{h}$, and the output $\Tilde{y}_{rnn}\in R^{\Tilde{T}}$. The $g(\cdot)$ is the rectified
linear unit (ReLu) activation function, which computes $f:x \mapsto max(0,x)$. \\
While the LSTM captures the context between links via message passing of computed internal cell states (capturing sequential dependencies), the CNN implicitly searches for direct local structures and interaction between the links as a feature extractor mechanism. The combination of CNN and LSTM allows us to enrich the feature space and expand the receptive field of trips over the links.

\subsection{Loss Function and Training}
Our goal is to estimate electric consumption at the link level while ensuring trip-energy accuracy is preserved, that is $\hat{y}^{i}_{trip}=\sum_{t=1}^{\Tilde{T}}\hat{y}_{t,i}$. A new trip can be a new combination of existing (adjacent) links and unknown (i.e previously unseen) to our model. Therefore, we require our model to be accurate on both link-level and trip-level consumption. Meanwhile, to further improve our model's generalization ability, we impose an additional cumulative link loss that requires the model to be precise on partial trips up to each link. Our loss is formulated as:  
\begin{equation}
\label{eqn:loss}
\begin{aligned}
    \mathcal{L}(\Theta)=  \frac{1}{N}\sum_{i=1}^{N}\Bigg( & \frac{1}{T_i}\sum_{l=1}^{T_i}\Big[\sum_{t=1}^{l}\big(y_{t,i}-f(\Theta, \mathbf{x}_{t,i})\big)\Big]^2\\
    &+\Big[\sum_{t=1}^{T_i}\big(y_{t,i}-f(\Theta,\mathbf{x}_{t,i})\big)\Big]^2 \\
    &+\frac{1}{T_i}\sum_{t=1}^{T_i}\big(y_{t,i}-f(\Theta,\mathbf{x}_{t,i})\big)^2 \Bigg)
\end{aligned}
\end{equation}
We search the optimal solution, $$\Theta^* = \argmin_{\Theta} \mathcal{L}(f(\Theta)
)$$ through back-propagation gradient descent, where,  $$C_l=\frac{1}{T_i}\big[\sum_{t=1}^{l}\big(y_{t,i}-f(\Theta, \mathbf{x}_{t,i})\big)\big]^2$$ evaluates the model's performance on the partial trip up to the $l$-th link. We sum $C_l$ over all $l$ to be the cumulative link loss, $$C_{cum}=\sum_{l=1}^{T_i}C_l$$ which evaluates the model's predictions on partial trips up to each link. The trip-level loss $$C_{trip}=\big[\sum_{t=1}^{T_i}\big(y_{t,i}-f(\Theta,\mathbf{x}_{t,i})\big)\big]^2=(\hat{y}^{i}_{trip}-y^{i}_{trip})^2$$ evaluates the model's prediction on the final consumption of the entire trip, and $$C_{link}=\frac{1}{T_i}\sum_{t=1}^{T_i}\big(y_{t,i}-f(\Theta,\mathbf{x}_{t,i})$$ measures the model's performance on each link. As mentioned in the previous section, we group trips based on their length without padding. Trips in the same batch share a common $T_i$, so the objective function \eqref{eqn:loss} is precise in training.  \\

\subsection{Architecture Details and Implementation}
The sequence modeling aspect of the model is wrapped in an encoder-decoder architecture \cite{cho2014learning}. This allows the model to learn whole trip representations using an RNN encoder block. The encoder summarizes the entire trip into a fixed-length vector $c$ representing the entire sequence. The decoder is a second RNN block that is trained to generate the electric consumption outputs at each link. In the decoder, $y_{link,i}$ and $h_{link,i}$ for link $i$ are both conditioned on $c$, but also on the previous and next link states and inputs, via the bidirectional passes. The encoder and decoder are jointly trained to minimize the loss function in equation (\ref{eqn:loss}). In our implementation, our encoder RNN is a deep LSTM with 2 layers, and 512 units at each layer. The decoder is a bidirectional LSTM of 256 units initialized with the encoder hidden and cell states. We stack the decoder with several fully connected neural network layers of size 128, 64, 32, and the last layer has 1 unit. As explained before, the neural network layers are time distributed; in other words, the same weights are applied to the encoder output for each link.

\begin{figure}[ht!]
\includegraphics[width=\linewidth]{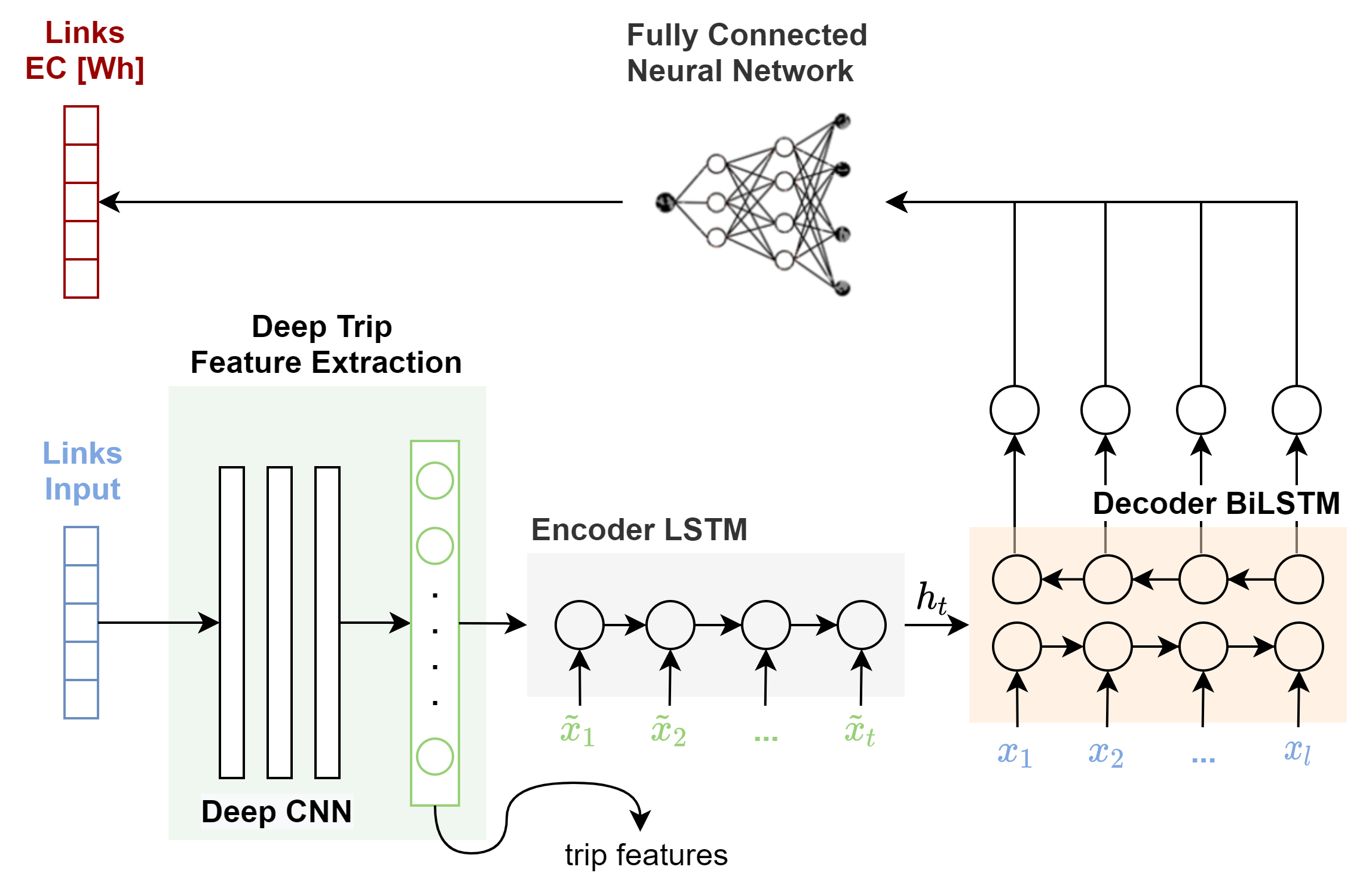}
\caption{General View of Model Architecture. Links for a trip constitute the input sequence that is fed to a deep convolution neural network for feature extraction. The processed sequence is encoded into a fixed latent vector $h_t$. The decoder recurrent network is bidirectional, and each link level output is conditioned on the encoder latent vector in addition to the current link input and the bidirectional hidden states of surrounding links. The decoder output is processed by a time-distributed fully connected layer to output an energy consumption estimate at the link level.}
\label{fig:architecture}
\end{figure}

The input data is pre-processed before being fed to the encoder-decoder, by a 3-layer deep convolution neural network with residual connections between the layers. Batch normalization and max-pooling are performed at each layer. Kernel size was fixed at 3, while the number of filters doubled at every layer, starting with 32 filters. The convolution neural network was leveraged as a feature extractor of important trip features while also shortening the sequence at every pass of a convolution block. See Figure \ref{fig:architecture} for an overview of the architecture.

Other implementation details:
\begin{itemize}
  \item We initialized all of the LSTM’s parameters with the uniform distribution. Non-recurrent weights were initialized by sampling from the isotropic zero-mean (white) Gaussian distribution.
  \item Optimization is performed with stochastic gradient descent without momentum. We employed a dynamic learning rate, initialized at 0.001 and decay proportional to $epoch^{-1/2}$.
  \item Due to bucketing of trips (as per Section \ref{sec:data}), the batch size is variable but capped at 128 trips. The bucketing speeds up processing because batches contain similar sequence length and no computation is wasted on the presence of potentially shorter sequences.
  \item Standard ReLu activation functions were used for convolutional and dense layers, and $tanh$ was used for recurrent layers.
  \item Excluding the convolution and recurrent layers, we used dropout regularization.
\end{itemize}

\subsection{Computational Setting}
For training and tuning purposes, we leveraged multiple graphical processing units (GPUs) spread across multiple machines with a multi-worker mirrored distributed training strategy. This strategy implements synchronous distributed training across multiple workers, each with potentially multiple GPUs. With the help of roughly 30 machines, each with at least 1 GPU, we were able to decrease the model training and tuning time for a dataset of this scale. Our hyperparameter search space contained different hyperparameters spanning different areas of model development including the different pre-processing approaches discussed in Section \ref{sec:data}, model selection and design, and learning rates. In the remaining sections, we present the results of the best trained model.

\section{Results and Analysis}
\subsection{Metrics}
We evaluate the model performance on root mean square error (RMSE) and mean absolute error (MAE) at the link and trip levels. However, there are some long trips in the dataset where the energy consumption accumulates to tens of thousands of Wh. In these cases, a 1\% error in prediction will lead to a large-magnitude RMSE and MAE. Therefore, we also evaluate our model on mean arctangent absolute percentage error (MAAPE). MAAPE is less sensitive to the standard mean absolute percentage error \cite{maape}, especially when $y^{i}_{\text{trip}}$ is small.\\
MAAPE is formulated as: 
$$\text{MAAPE}= \frac{1}{N}\sum_{i=1}^{N}\arctan \Big|\frac{y^{i}_{\text{trip}}- \hat{y}^{i}_{\text{trip}}}{y^{i}_{\text{trip}}}\Big|$$
It averages the AAPE, $\arctan \Big|\frac{y^{i}_{\text{trip}}- \hat{y}^{i}_{\text{trip}}}{y^{i}_{\text{trip}}}\Big|$ where each absolute percentage error is bounded by $\frac{\pi}{2}$ in MAAPE, so the value of MAAPE will not be significantly affected by a small number of outliers when the sample size is large. Arctangent absolute percentage error also approximates absolute percentage error well for small errors. From a Taylor series of $\arctan{x}$, we see that $|\arctan{x}-x|=\mathcal{O}(\epsilon^3)$ for $x \in [0,\epsilon]$. As a result, MAAPE can give us a better idea of the overall performance of the model when there are outliers in errors.

\subsection{Residual Analysis}
\subsubsection{Overall Performance}
We measure the error by magnitude in Table \ref{tab:magnitude}. The model has a fairly low overall MAE and RMSE on electric consumption predictions when compared to the range of the true values from the test set statistics shown. Trip-level performance results represent residual values over an entire trip (i.e., summed over all the links in trip). The MAAPE at trip level is less than 3\%. However on the link level, the MAAPE is close to 14\%, suggesting that the model's performance on individual links decreases when isolated. This phenomenon is due to our loss function choice on which the model was optimized, favoring lower error as it accumulates along a trip; in other words, accurate estimation of total battery energy usage is prioritized to capture SOC depletion and charging needs.

\begin{table}[H]
\begin{subtable}{\linewidth}
\centering
\centering
\begin{tabular}{c|cc}
\textbf{}                            & \multicolumn{2}{c}{\textbf{Test Set Model Performance Metrics}} \\ \cline{2-3} 
\multirow{3}{*}{\textbf{Trip level}} & \multicolumn{1}{c|}{\textbf{RMSE}}           & 96.7 Wh          \\
                                     & \multicolumn{1}{c|}{\textbf{MAE}}            & 55.3 Wh          \\
                                     & \multicolumn{1}{c|}{\textbf{MAAPE}}           & 0.024            \\ \hline
\multirow{3}{*}{\textbf{Link level}} & \multicolumn{1}{c|}{\textbf{RMSE}}           & 22.9   Wh          \\
                                     & \multicolumn{1}{c|}{\textbf{MAE}}            & 14.5 Wh              \\
                                     & \multicolumn{1}{c|}{\textbf{MAAPE}}           & 0.142          
\end{tabular}
\caption{Test set mean absolute error (MAE) \& Root Mean Squared Error (RMSE) \& MAAPE of trip and link-level electric consumption.}

\vspace{0.5cm}

\resizebox{\textwidth}{!}{%
\begin{tabular}{c|cccllc}
                                     & \multicolumn{6}{c}{\textbf{Test Set Electric Consumption Stats {[}Wh{]}}}                                                                                     \\ \cline{2-7} 
                                     & \textbf{Mean}           & \textbf{Std}            & \textbf{Min}        & \textbf{50\%}         & \textbf{75\%}           & \textbf{Max}             \\ \cline{2-7} 
\multirow{2}{*}{\textbf{Trip level}} & \multirow{2}{*}{3391.1} & \multirow{2}{*}{5078.8} & \multirow{2}{*}{14} & \multirow{2}{*}{1789} & \multirow{2}{*}{3951.2} & \multirow{2}{*}{92159.4} \\
                                     &                         &                         &                     &                       &                         &                          \\ \hline
\multirow{2}{*}{\textbf{Link level}}                  & \multirow{2}{*}{180}                     & \multirow{2}{*}{286.4}                   & \multirow{2}{*}{-754}                & \multirow{2}{*}{108.9}                 & \multirow{2}{*}{213.2}                   & \multirow{2}{*}{25476.9}     \\
&                         &                         &                     &                       &                         &                          \\
\end{tabular}%
}
\caption{Test Set Electric Consumption Distribution Statistics}
\end{subtable}%

\caption{Model Performance metrics compared with electric consumption distribution in the test dataset.}
\label{tab:magnitude}
\end{table}

\begin{figure}[ht!]
\includegraphics[width=\linewidth]{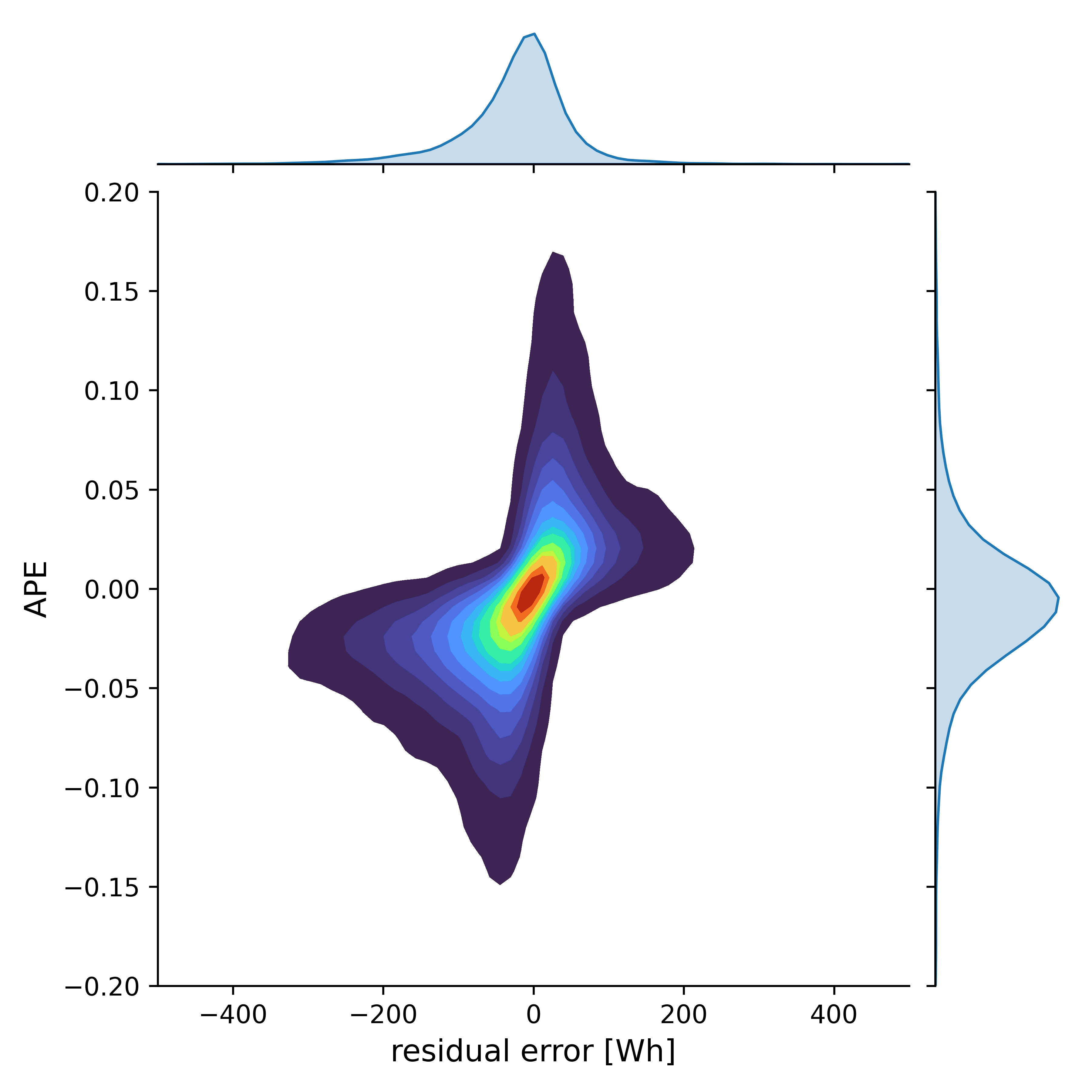}
\caption{Trip-level joint density plot of APE, calculated akin to AAPE without the absolute value component, \& residual error $r = y_{trip} -\hat{y}_{trip}$ of true consumption versus predicted.}
\label{fig:density}
\end{figure}

Figure \ref{fig:density} shows a joint density plot of arctangent percentage error (APE) $\text{APE}= \arctan \frac{y^{i}_{\text{trip}}- \hat{y}^{i}_{\text{trip}}}{y^{i}_{\text{trip}}}$ with the raw residual error $r = y_{trip} -\hat{y}_{trip}$, which sheds light on the model fitting behavior. In Section \ref{tab:magnitude} we observed very low error rates at the trip level despite having higher errors at the link level, even though a trip is constructed from a sequence of links. The density plot suggests that higher APE concentrates around low residuals (i.e., the true consumption is likely to be very small). In other words, small and short links that manifest very low consumption values do not impact model performance, because they contribute little to the overall result. We will see later that those represent short and brief transitional links. We also note that the marginal distribution over APE and the residuals are normal and centered around zero.

\begin{figure}[ht!]
\includegraphics[width=\linewidth]{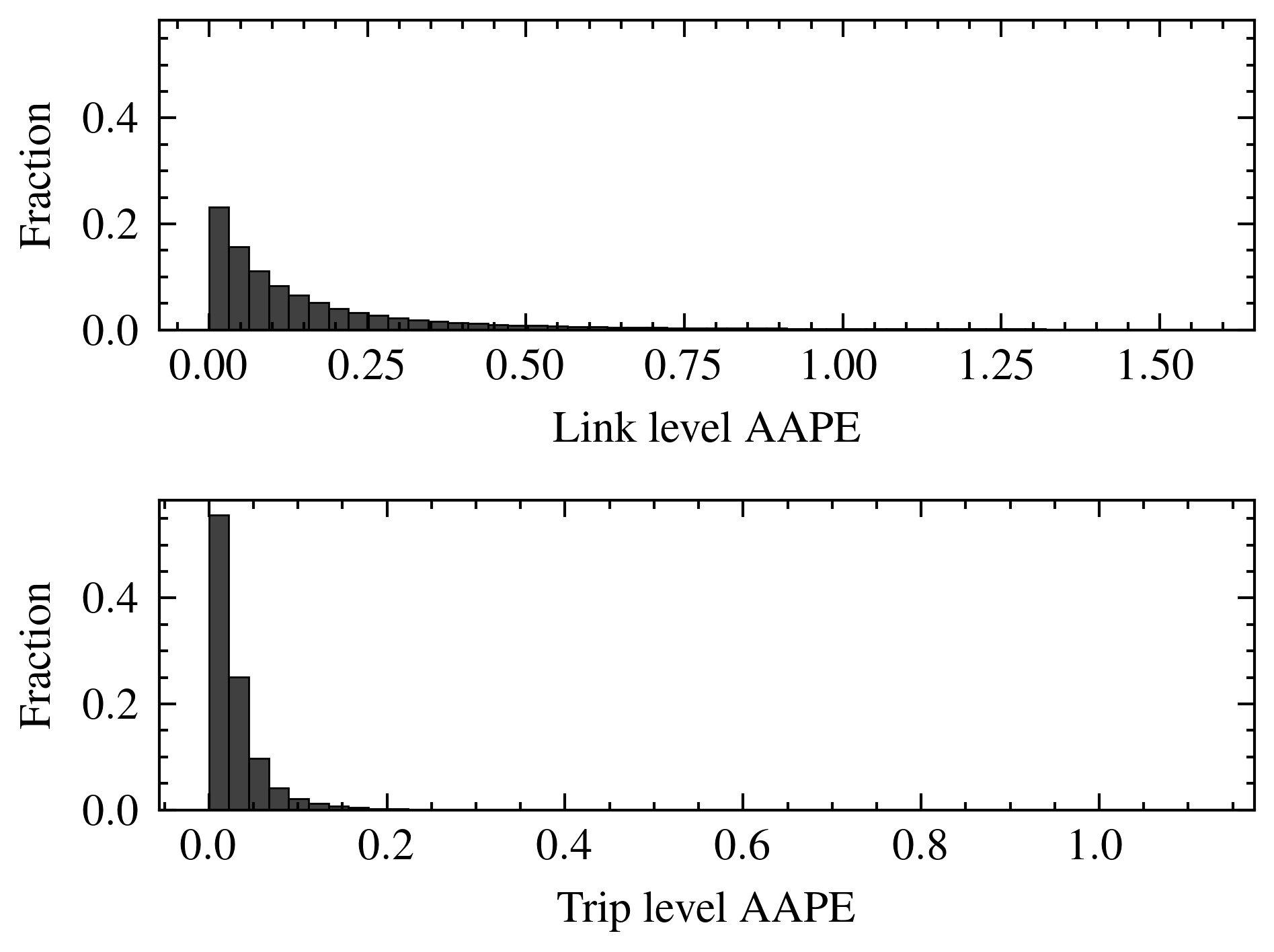}
\caption{AAPE distribution over links and trips}
\label{fig:hist}
\end{figure}

We compute the AAPE of electric consumption over all links and trips and list the percentiles of those errors in Table \ref{tab:percentiles} for more a granular view. For more than half of the trips, the electric consumption predictions of the model are off by less than 1.9\% and 9.4\% on the trip and link level, respectively. On the trip level, predictions remain contained within 6.5\% error for more than 90\% of the trips, although they can go up to \%50 error on the link level. In fact, Figure \ref{fig:hist} reveals that AAPE distribution over links is heavily right skewed with a rather fat tail.

\begin{table}[ht!]
\centering
    \begin{tabular}{c c c c c c} 
        \hline
        Consumption  &25th & 50th & 75th & 90th & 95th\\ 
        \hline 
        Trip   & 0.008 & 0.019 & 0.037 & 0.065 & 0.092\\
        Link   & 0.034 & 0.094 & 0.228 & 0.493 & 0.778 \\ 
        \hline 
    \end{tabular}
\caption{Trip and link level AAPE percentiles of electric consumption.}
\label{tab:percentiles}
\end{table}

\begin{figure}[ht!]
\includegraphics[width=\linewidth]{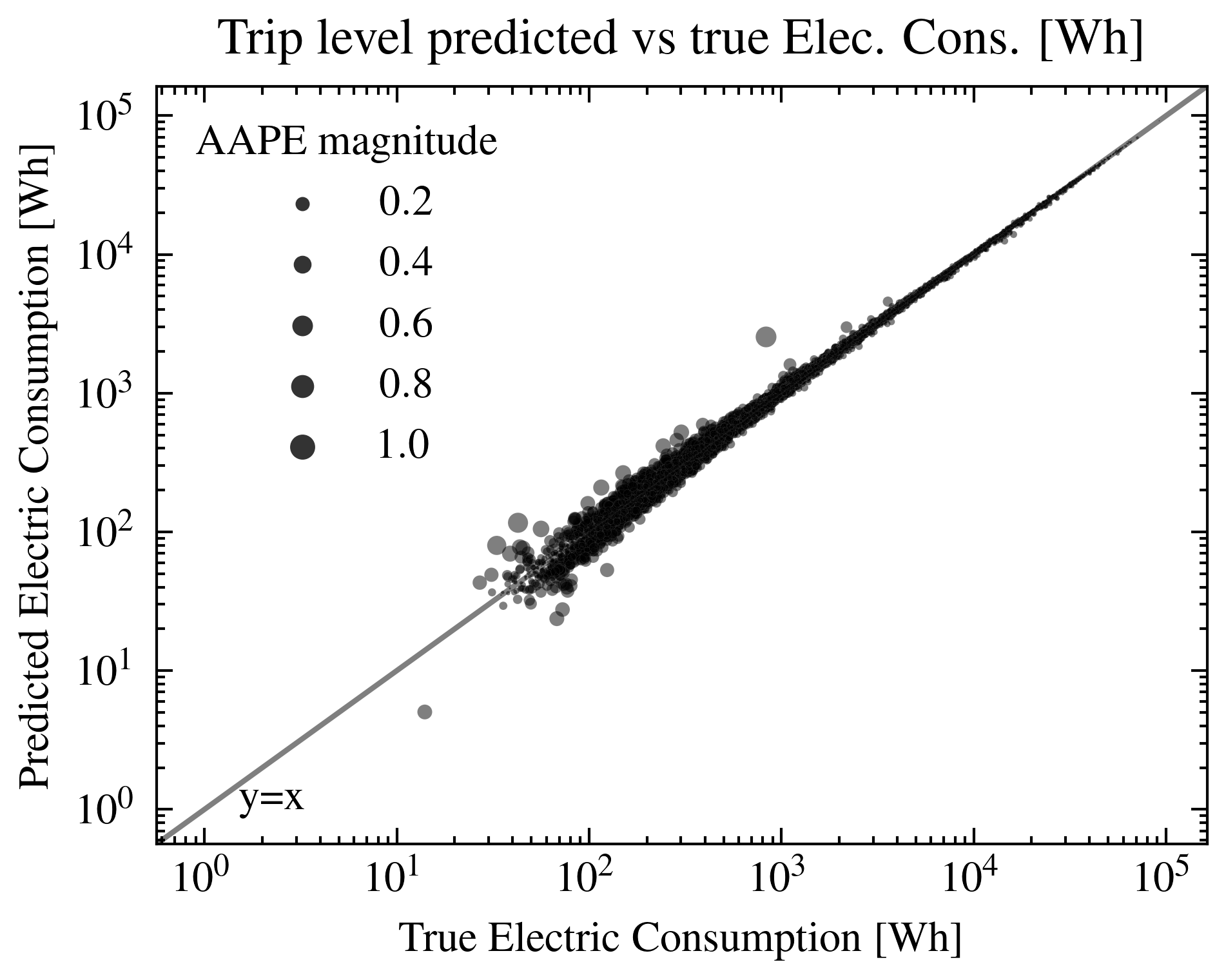}
\caption{Scatter plot of predicted electric energy consumption on the test dataset versus true consumption.}
\label{fig:scatter}
\end{figure}

A final look at overall model performance can be seen in Figure \ref{fig:scatter}, where the prediction electric consumption point for all the trips in the test dataset are plotted against the true consumption on a log scale. The model presents a very high $R^2=0.99$.

\subsubsection{Detailed Example Result}
In this subsection, we go over the details of two predicted trip results randomly sampled from the test set: Trip \#56 and Trip \#110.

\begin{figure}[!t]
  \begin{subfigure}[b]{\linewidth}
    \includegraphics[width=\textwidth]{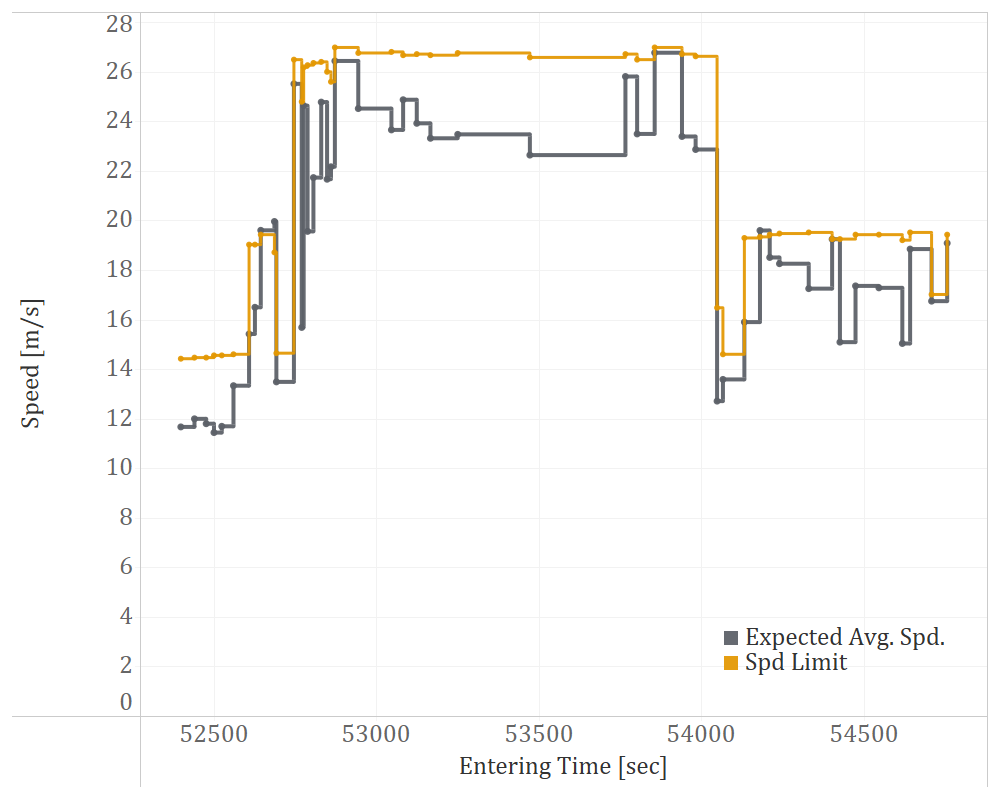}
    \caption{Route profile: Output of POLARIS, Input of Model.}
    \label{fig:example_spd_t56}
  \end{subfigure}
  \hfill
  \begin{subfigure}[b]{\linewidth}
    \includegraphics[width=\textwidth]{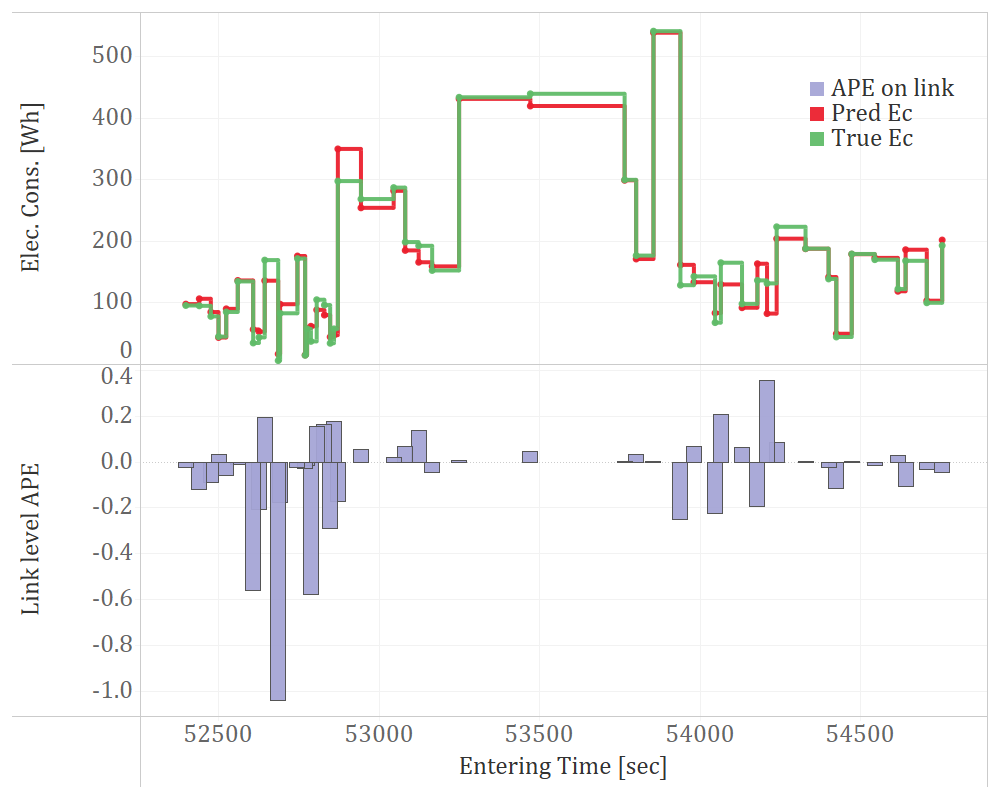}
    \caption{Link level predicted \& true energy (top). Link level APE (bottom).}
    \label{fig:example_link_t56}
  \end{subfigure}
  \hfill
  \begin{subfigure}[b]{\linewidth}
    \includegraphics[width=\textwidth]{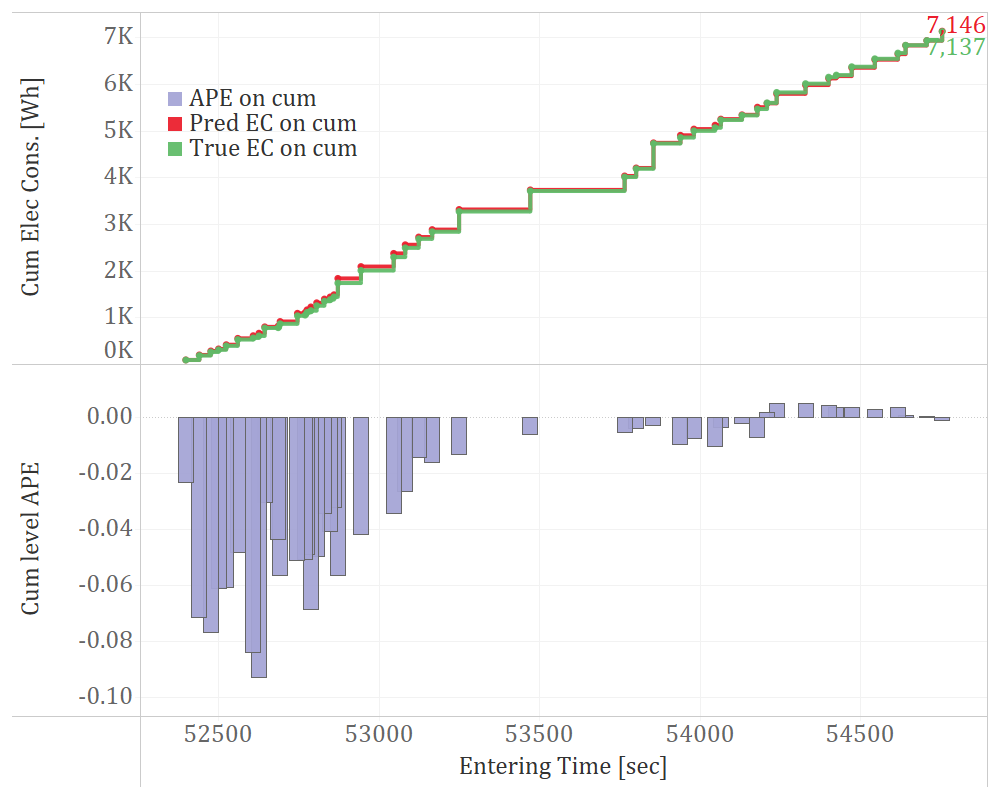}
    \caption{Cumulative predicted \& true energy (top). APE (bottom)}
    \label{fig:example_cum_t56}
  \end{subfigure}
  \caption{Trip\#56: Analysis of energy predictions against True.}
\end{figure}

\begin{figure}[!t]
  \begin{subfigure}[b]{\linewidth}
    \includegraphics[width=\textwidth]{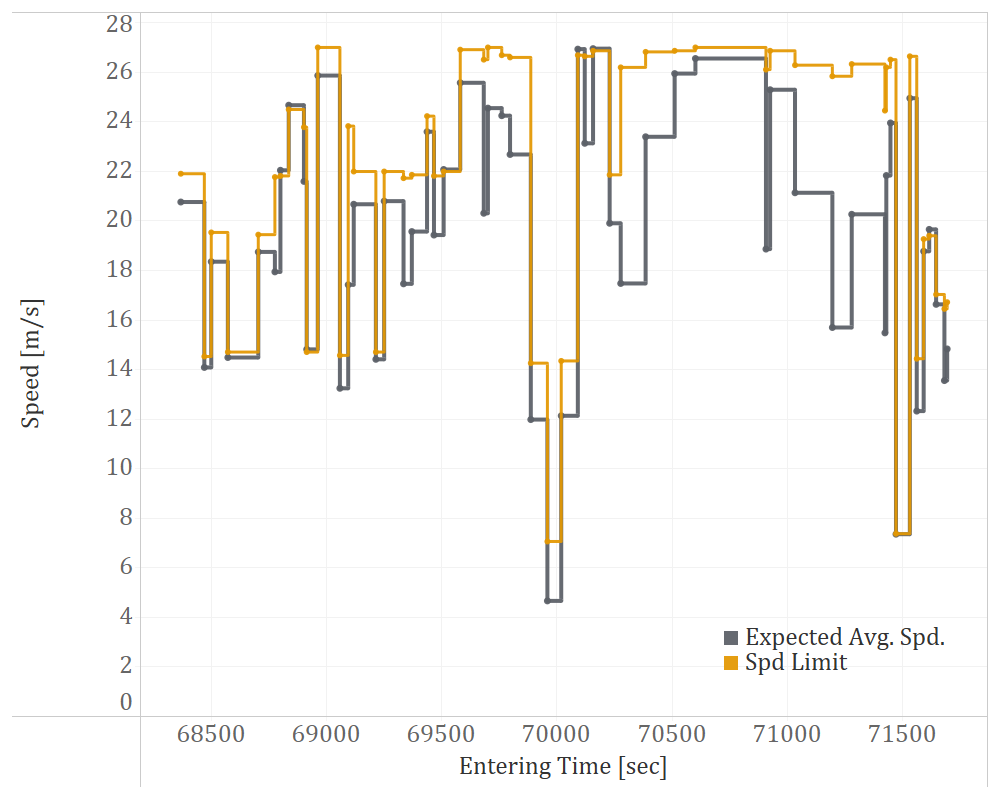}
    \caption{Route profile: Output of POLARIS, Input of Model.}
    \label{fig:example_spd_t110}
  \end{subfigure}
  \hfill
  \begin{subfigure}[b]{\linewidth}
    \includegraphics[width=\textwidth]{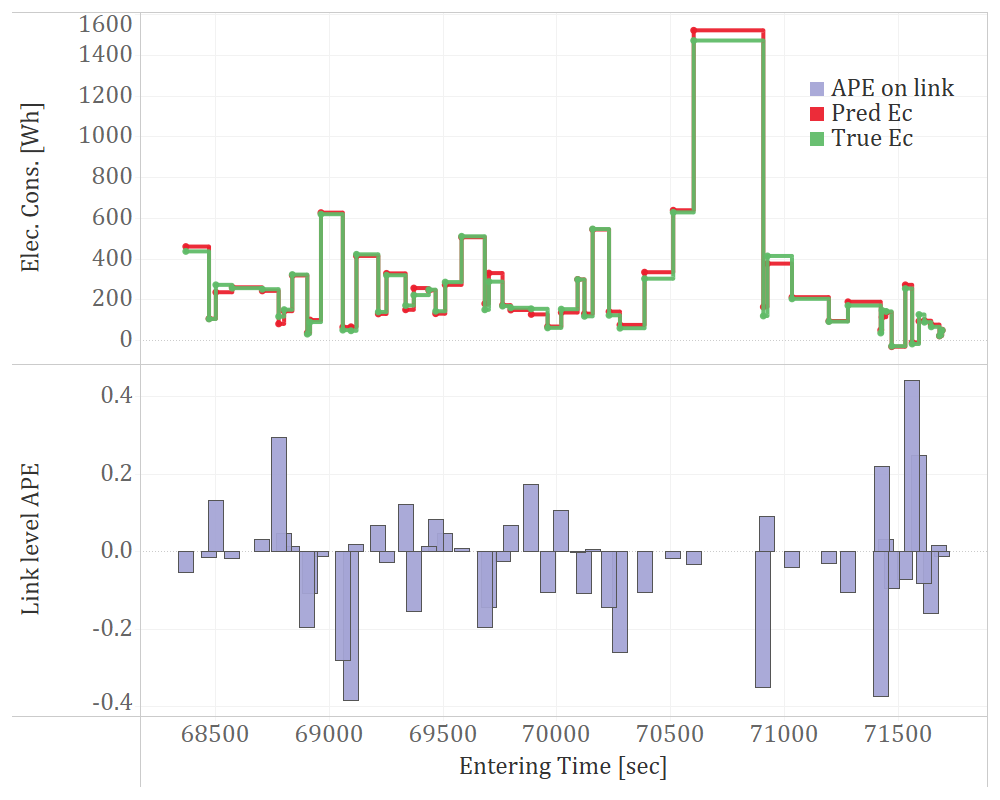}
    \caption{Link level predicted \& true energy (top). Link level APE (bottom).}
    \label{fig:example_link_t110}
  \end{subfigure}
  \hfill
  \begin{subfigure}[b]{\linewidth}
    \includegraphics[width=\textwidth]{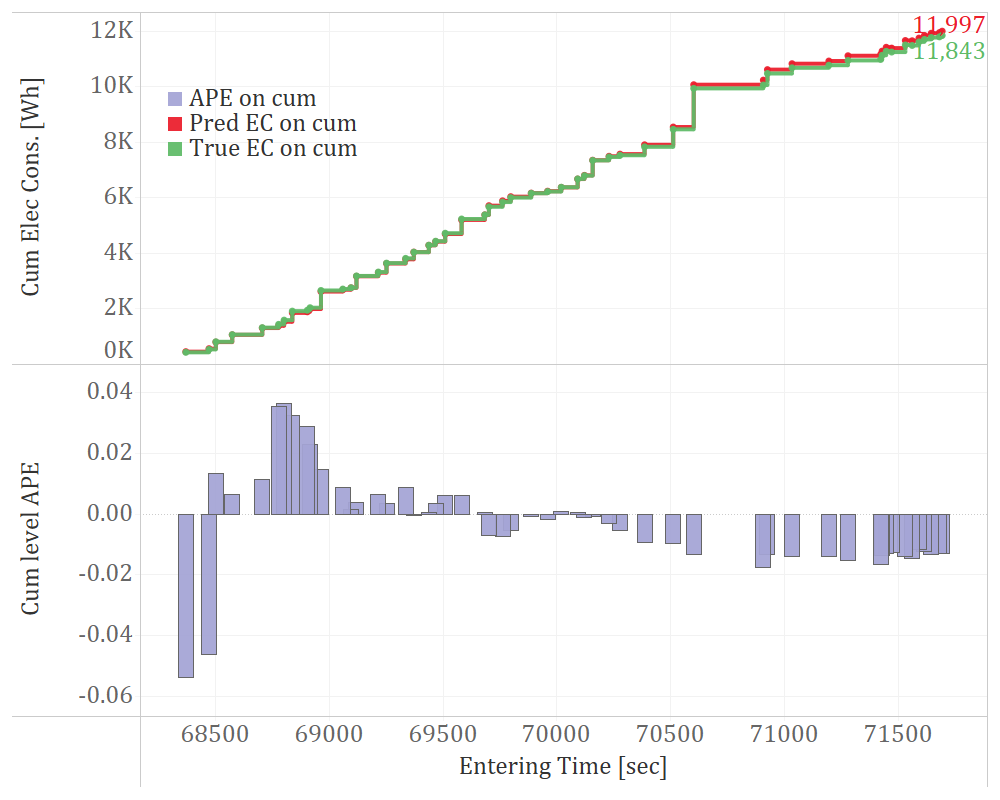}
    \caption{Cumulative predicted \& true energy (top). APE (bottom)}
    \label{fig:example_cum_t110}
  \end{subfigure}
  \caption{Trip\#110: Analysis of energy predictions against True.}
\end{figure}

Figure \ref{fig:example_spd_t56} and \ref{fig:example_spd_t110} show the high-level route profile outputs from simulated agents in POLARIS. This high-level routing information along with the vehicle feature information is provided to the model as input, as described in Section \ref{sec:DLS}. Figures \ref{fig:example_link_t56} and \ref{fig:example_link_t110} show the link-by-link predicted electric consumed output in Wh along with the true energy from the test set. We observe precise link-level estimates for the most part, except when the vehicle transitions to a very short, low consumption link. In Figures \ref{fig:example_cum_t56} and \ref{fig:example_cum_t110}, we provide the cumulative energy consumed over the trip (top) and the APE computed at each accumulation step (i.e., by the end of the trip the last bar displays the trip-level energy error). Based on the (b) and (c) APE bar charts of both trips we recognize the interplay between positive and negative link-level APE and its impact on the cumulative error decreasing trend, as per the design of the model and its loss function. On both trips, the final APE is small: -0.12\% and -1.3\% for Trips \#56 and \#110, respectively.

\subsubsection{Network-Wide Application}
\begin{figure}[ht!]
\includegraphics[width=\linewidth]{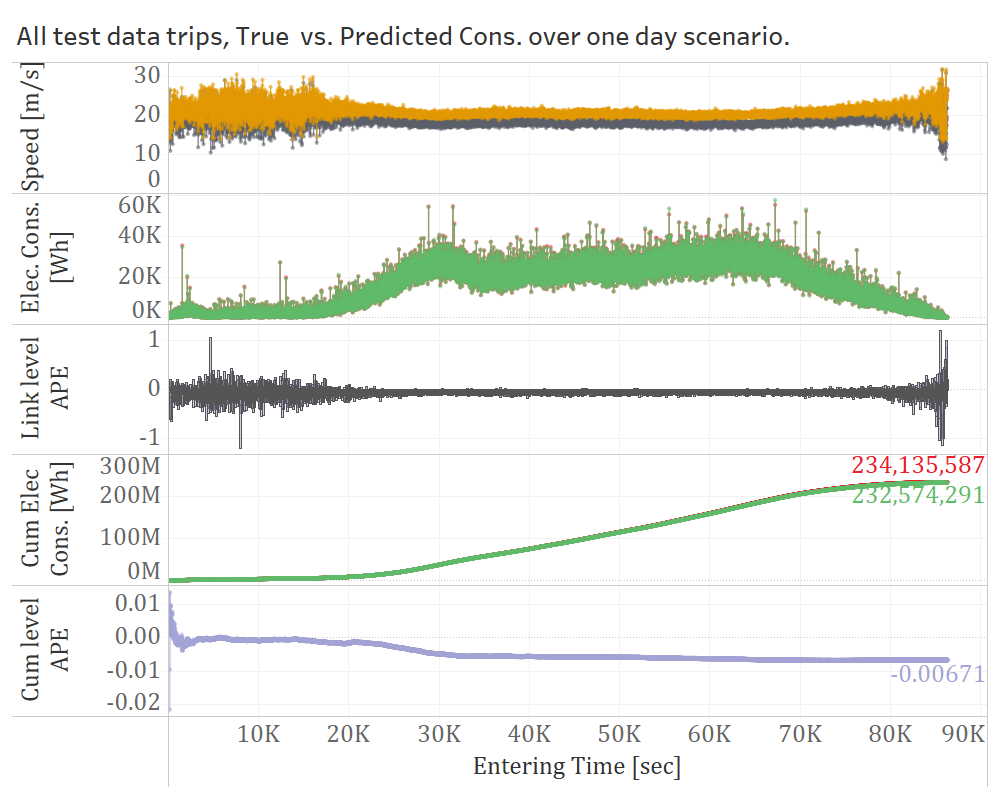}
\caption{All trips in test data.}
\label{fig:example_all}
\end{figure}

Figure \ref{fig:example_all} provides an estimate of the predicted energy and its accuracy over the entirety of the test dataset. This can represent, for example, the complete mobility of agents within the region. The total energy consumed at the macroscopic level can be predicted and compared against the true energy outcomes aggregated from microscopic simulation results. In fact, the figure shows that the cumulative predicted energy is in the order of 234 MWh while the true energy calculated from the test set data is 232 MWh, which gives a network-wide energy consumption error of 0.6\%.

\section{Application to Charging Behavior}
The model developed here through the processes described above provides a robust way to make predictions on energy consumption at the link level based on vehicle and route characteristics. The end goal of this exercise is to abstract the process carried out in estimating this model, that is the integrated process of stitching data from POLARIS, SVTrip, and Autonomie to produce robust energy consumption estimates. While the model was estimated rigorously through the TensorFlow framework using several GPUs to perform parameter updates in the fully-connected neural net, predictions using this framework can be computational expensive when analyzing several million vehicles in large regional simulations. The model developed was transferred onto the TensorFlowLite (TFLite) framework, which is a light-weight deployable version of TensorFlow. TFLite performs necessary optimizations on neural network structure and parameter precision to minimize prediction time while trying to retain a similar level of accuracy as the input model.

With POLARIS running on a high-performance C++ framework, the C++ API for TFLite was used to integrate the transferred model for use within the POLARIS simulation workflow. This API allows for quick allocating of non-contiguous memory and prepares the environment for model predictions. A tensor of the same length as used for estimation described above is the only required input to carry out prediction at the link level. By deploying this model at the link level, a continuous update of vehicle SOC becomes possible allowing travelers to make decisions on charging the battery throughout the day. Since TFLite abstracts parameter precision and performs other optimizations for quick prediction turnaround, the existing high-speed runtimes achieved in POLARIS are maintained with minimal changes in overall run time. This allows to continue modeling the travel demand for large regions, now including robust EV consumption tracking and charging events. Each prediction of battery consumption at the link level takes $\leq 50$ms per call.

\begin{figure}[ht!]
\centering
\includegraphics[width=3.4in, clip]{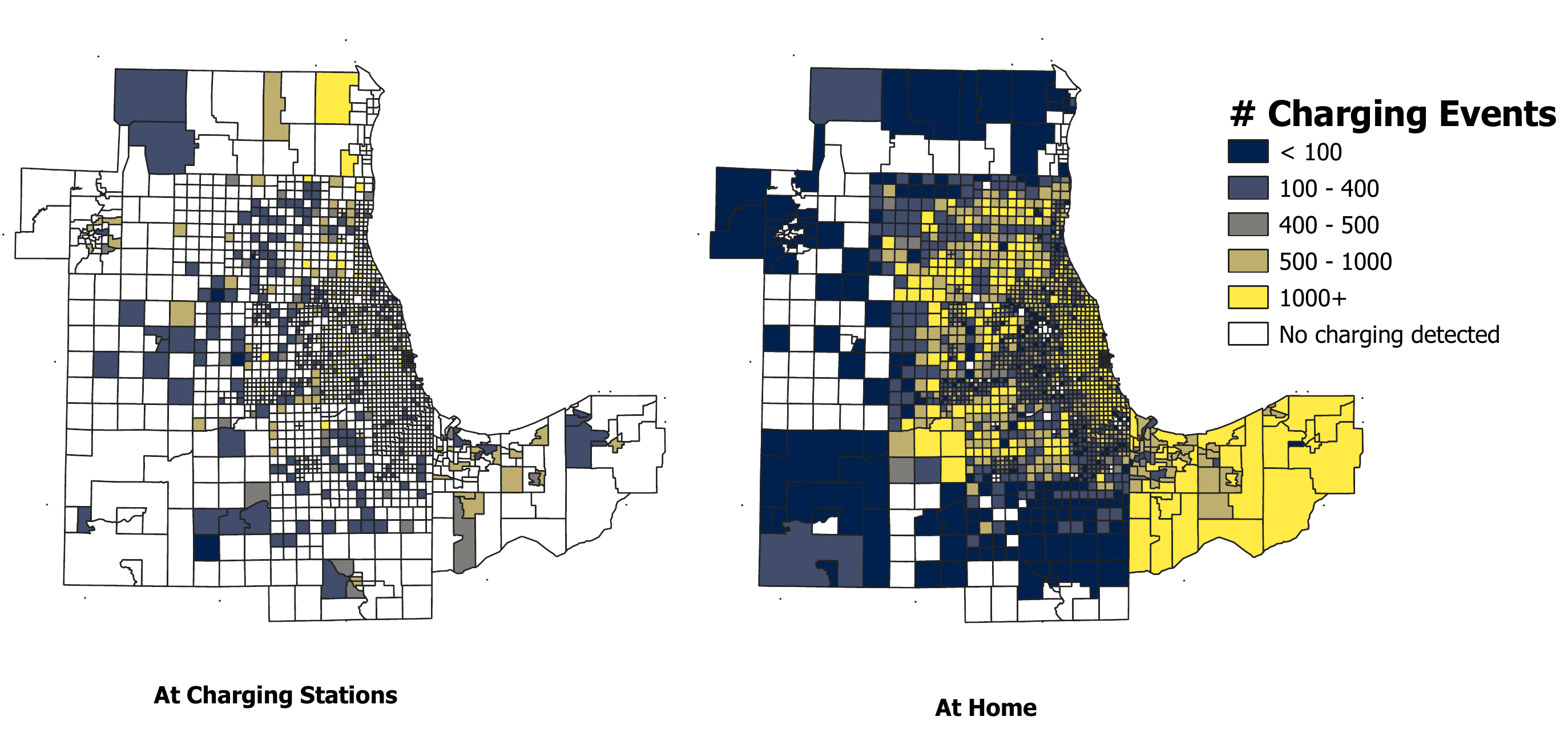}
\caption{Charging Events in Chicago at Home and at Charging Stations
}
\label{fig:ChargingEvents}
\end{figure}

\begin{figure}[ht!]
\centering
\includegraphics[width=3.4in, clip]{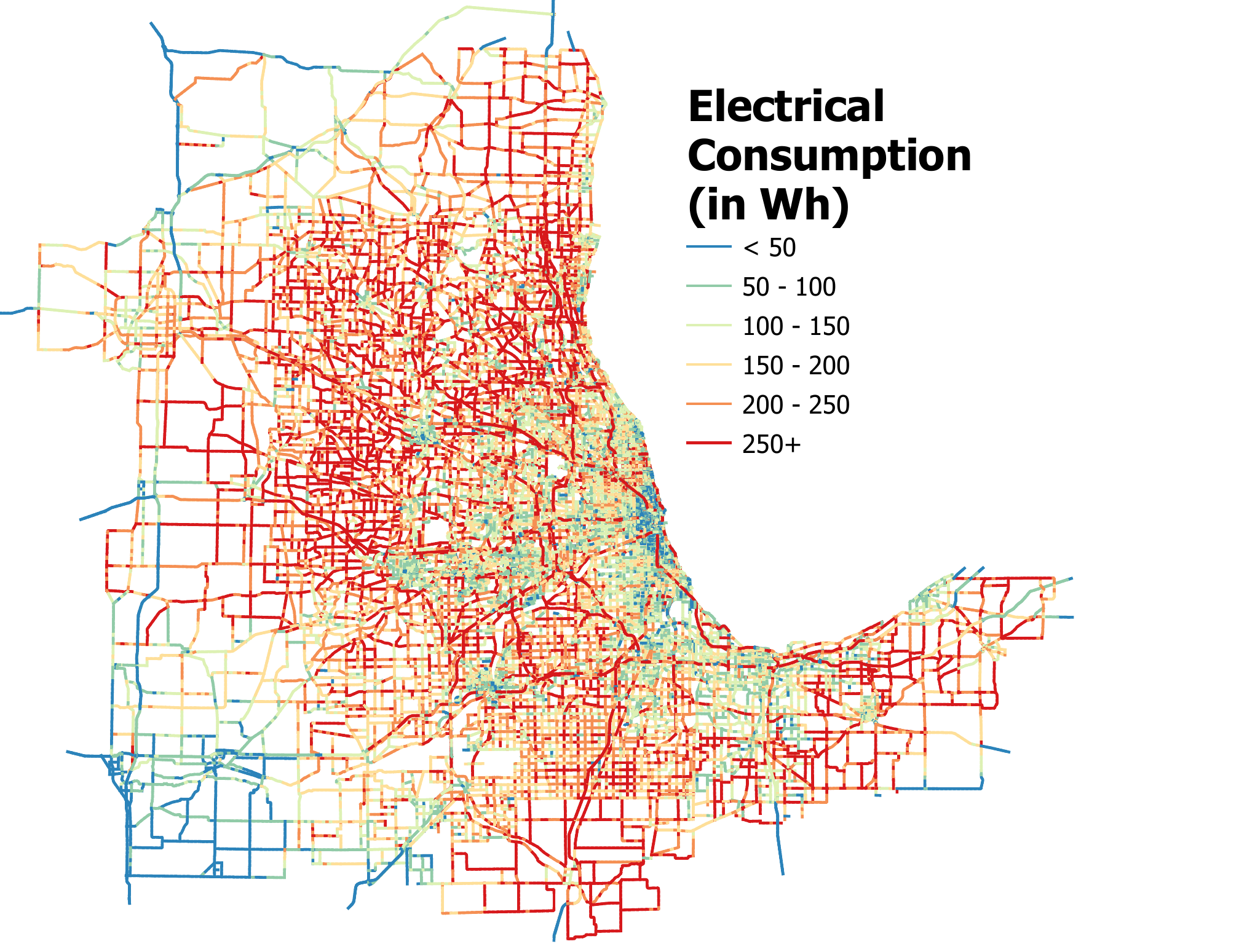}
\caption{Predicted Link Consumption across Greater Chicago via POLARIS
}
\label{fig:LinkConsumption}
\end{figure}

An example application of this process is visualized in Figures \ref{fig:ChargingEvents} and \ref{fig:LinkConsumption} for the case of the Greater Chicago region. Figure \ref{fig:ChargingEvents} shows the spatial demand structure of EV charging events throughout the region grouped by underlying traffic analysis zones (typically used by local Metropolitan Planning Organizations, or MPOs). Figure \ref{fig:LinkConsumption} shows average energy use by link in the region by averaging EV consumption of all vehicles passing through these links. Moving forward, a more detailed scenario setup will be studied to understand how varying electric vehicle penetration rates in a region and location and available of EV charging station impacts vehicle use and charging demands.

\section{Discussion}
Our work connects with a broad literature on electric consumption prediction of EVs, where the proposed approaches are categorized into vehicle model-driven methods and data-driven methods. The lack of, and difficulty in acquiring, microscopic data are major factors that hindered research advances until massive real-world driving data is available to predict the energy consumption of EVs with machine-learning algorithms. In most common settings, estimates are made on the basis of high-fidelity speed profiles and leveraging machine learning (\cite{YAO2019276}, \cite{de_cauwer_data-driven_2017}, \cite{MODI2020454}, \cite{ZHANG2020115408}, \cite{GENIKOMSAKIS201798}). Some studies have already explored the use of CNN to estimate the energy consumption of EVs \cite{MODI2020454}, while others use classical statistical methods \cite{LOPEZ2020124188}, but again in the presence of precise stochastic speed profiles. In the meantime, Al-Wreikat et al. \cite{ALWREIKAT2021117096} have carried out statistical analysis on how driving behavior and conditions affect the energy consumption of EVs, \cite{fiori_effect_2019} and \cite{XIE2020115081} pointed out that energy consumption is highly related to average traffic speed. \cite{qi_data-driven_2016} and \cite{QI201836} are some of the rare studies that employ such constraints to the modeling, by using average link-level speed information only. \cite{YI2018344} proposed a data-driven density estimation method that accounts for trip average speed, and \cite{BASSO2019141} attempted to derive explicit equations via thorough mathematical modeling, then took a machine-learning approach in \cite{BASSO202124}. All such studies are motivated by the need to engage transportation network-level use-case applications (e.g., route planning, eco-routing, charging behavior). In general, we argue that it is challenging to estimate energy consumption on the basis of actual vehicle speed information because driving behaviors and traffic conditions are uncertain and heterogeneous. A predictive model trained only on macroscopic route input data but fine-tuned to equal microscopic results seems to be a more realistic approach. As we showed by designing a very large experiment using energy data generation from high-fidelity simulations and by the subsequent masking of microscopic information, a deep-learning approach can overcome latent information and retrieve accurate aggregated energy consumption values.

In this work, environment-related factors have been left out. In future work, the inclusion of ambient temperature, weather conditions and road grade \cite{LIU201774} at the macroscopic level will be of interest to study within this modeling framework.

\section*{Acknowledgments}
The work described was sponsored by the U.S. Department of Energy (DOE) Vehicle Technologies Office (VTO) under the Systems and Modeling for Accelerated Research in Transportation (SMART) Mobility Laboratory Consortium, an initiative of the Energy Efficient Mobility Systems (EEMS) Program. The following DOE Office of Energy Efficiency and Renewable Energy (EERE) managers played important roles in establishing the project concept, advancing implementation, and providing ongoing guidance: Erin Boyd, Danielle Chou, Heather Croteau, Prasad Gupte, and David Anderson. The submitted manuscript was created by UChicago Argonne, LLC, Operator of Argonne National Laboratory (Argonne). Argonne, a U.S. Department of Energy Office of Science laboratory, is operated under Contract No. DE- AC02-06CH11357. The U.S. Government retains for itself, and others acting on its behalf, a paid-up nonexclusive, irrevocable worldwide license in said article to reproduce, prepare derivative works, distribute copies to the public, and perform publicly and display publicly, by or on behalf of the Government.\par

\bibliography{bibtex/IEEEabrv.bib,bibtex/reference.bib}{}

\begin{thebibliography}{10}
\providecommand{\url}[1]{#1}
\csname url@samestyle\endcsname
\providecommand{\newblock}{\relax}
\providecommand{\bibinfo}[2]{#2}
\providecommand{\BIBentrySTDinterwordspacing}{\spaceskip=0pt\relax}
\providecommand{\BIBentryALTinterwordstretchfactor}{4}
\providecommand{\BIBentryALTinterwordspacing}{\spaceskip=\fontdimen2\font plus
\BIBentryALTinterwordstretchfactor\fontdimen3\font minus
  \fontdimen4\font\relax}
\providecommand{\BIBforeignlanguage}[2]{{%
\expandafter\ifx\csname l@#1\endcsname\relax
\typeout{** WARNING: IEEEtran.bst: No hyphenation pattern has been}%
\typeout{** loaded for the language `#1'. Using the pattern for}%
\typeout{** the default language instead.}%
\else
\language=\csname l@#1\endcsname
\fi
#2}}
\providecommand{\BIBdecl}{\relax}
\BIBdecl

\bibitem{moawad_assessment_2016}
\BIBentryALTinterwordspacing
A.~Moawad, N.~Kim, N.~Shidore, and A.~Rousseau,
  ``\BIBforeignlanguage{English}{Assessment of {Vehicle} {Sizing}, {Energy}
  {Consumption} and {Cost} {Through} {Large} {Scale} {Simulation} of {Advanced}
  {Vehicle} {Technologies}},'' Argonne National Lab. Argonne, IL (United
  States), Tech. Rep. ANL/ESD-15/28, Jan. 2016. [Online]. Available:
  \url{https://www.osti.gov/biblio/1245199}
\BIBentrySTDinterwordspacing

\bibitem{islam_extensive_2018}
\BIBentryALTinterwordspacing
E.~Islam, A.~Moawad, N.~Kim, and A.~Rousseau, ``\BIBforeignlanguage{English}{An
  {Extensive} {Study} on {Sizing}, {Energy} {Consumption}, and {Cost} of
  {Advanced} {Vehicle} {Technologies}},'' Argonne National Lab. Argonne, IL
  (United States), Tech. Rep. ANL/ESD-17/17, Oct. 2018. [Online]. Available:
  \url{https://www.osti.gov/biblio/1463258-extensive-study-sizing-energy-consumption-cost-advanced-vehicle-technologies}
\BIBentrySTDinterwordspacing

\bibitem{islam_energy_2020}
\BIBentryALTinterwordspacing
E.~S. Islam, A.~Moawad, N.~Kim, and A.~Rousseau,
  ``\BIBforeignlanguage{English}{Energy {Consumption} and {Cost} {Reduction} of
  {Future} {Light}-{Duty} {Vehicles} through {Advanced} {Vehicle}
  {Technologies}: {A} {Modeling} {Simulation} {Study} {Through} 2050},''
  Argonne National Lab. (ANL), Argonne, IL (United States), Tech. Rep.
  ANL/ESD-19/10, Jun. 2020. [Online]. Available:
  \url{https://www.osti.gov/biblio/1647165-energy-consumption-cost-reduction-future-light-duty-vehicles-through-advanced-vehicle-technologies-modeling-simulation-study-through}
\BIBentrySTDinterwordspacing

\bibitem{islam_energy_2021}
E.~Islam, R.~Vijayagopal, A.~Moawad, N.~Kim, B.~Dupont, D.~Nieto~Prada, and
  A.~Rousseau, ``A detailed vehicle modeling \& simulation study quantifying
  energy consumption and cost reduction of advanced vehicle technologies
  through 2050,'' Argonne National Laboratory, Tech. Rep. ANL/ESD-21/10,, Oct
  2021, ,.

\bibitem{auld2016polaris}
J.~Auld, M.~Hope, H.~Ley, V.~Sokolov, B.~Xu, and K.~Zhang, ``Polaris:
  Agent-based modeling framework development and implementation for integrated
  travel demand and network and operations simulations,'' \emph{Transportation
  Research Part C: Emerging Technologies}, vol.~64, pp. 101--116, 2016.

\bibitem{freyermuth_energy_2019}
V.~Freyermuth, J.~Auld, D.~Karbowski, A.~Moawad, S.~Pagerit, and A.~Rousseau,
  ``Energy {Prediction} of the {Chicago} {Metropolitan} {Area} {Using}
  {Distributed} {Transportation} {MBSE} {Framework},'' in \emph{2019 {IEEE}
  {Vehicle} {Power} and {Propulsion} {Conference} ({VPPC})}, Oct. 2019, pp.
  1--7, iSSN: 1938-8756.

\bibitem{freyermuth_powertrain_2020}
------, ``Powertrain distribution based on total cost of ownership for
  privately owned vehicles and {TNC} in the {Chicago} metropolitan area,'' in
  \emph{2020 {IEEE} {Vehicle} {Power} and {Propulsion} {Conference} ({VPPC})},
  Nov. 2020, pp. 1--6, iSSN: 1938-8756.

\bibitem{gurumurthy2020integrating}
K.~M. Gurumurthy, F.~de~Souza, A.~Enam, and J.~Auld, ``Integrating supply and
  demand perspectives for a large-scale simulation of shared autonomous
  vehicles,'' \emph{Transportation Research Record}, vol. 2674, no.~7, pp.
  181--192, 2020.

\bibitem{karbowski_trip_2014}
\BIBentryALTinterwordspacing
D.~Karbowski, A.~Rousseau, V.~Smis-Michel, and V.~Vermeulen,
  ``\BIBforeignlanguage{English}{Trip {Prediction} {Using} {GIS} for {Vehicle}
  {Energy} {Efficiency}},'' Argonne National Lab. (ANL), Argonne, IL (United
  States), Tech. Rep., Jan. 2014. [Online]. Available:
  \url{https://www.osti.gov/biblio/1494826-trip-prediction-using-gis-vehicle-energy-efficiency}
\BIBentrySTDinterwordspacing

\bibitem{XAI_moawad}
A.~Moawad, E.~Islam, N.~Kim, R.~Vijayagopal, A.~Rousseau, and W.~B. Wu,
  ``Explainable ai for a no-teardown vehicle component cost estimation: A
  top-down approach,'' \emph{IEEE Transactions on Artificial Intelligence},
  vol.~2, no.~2, pp. 185--199, 2021.

\bibitem{lecun_deep_2015}
\BIBentryALTinterwordspacing
Y.~LeCun, Y.~Bengio, and G.~Hinton, ``\BIBforeignlanguage{en}{Deep learning},''
  \emph{\BIBforeignlanguage{en}{Nature}}, vol. 521, no. 7553, pp. 436--444, May
  2015, bandiera\_abtest: a Cg\_type: Nature Research Journals Number: 7553
  Primary\_atype: Reviews Publisher: Nature Publishing Group Subject\_term:
  Computer science;Mathematics and computing Subject\_term\_id:
  computer-science;mathematics-and-computing. [Online]. Available:
  \url{https://www.nature.com/articles/nature14539}
\BIBentrySTDinterwordspacing

\bibitem{NIPS2012_c399862d}
\BIBentryALTinterwordspacing
A.~Krizhevsky, I.~Sutskever, and G.~E. Hinton, ``Imagenet classification with
  deep convolutional neural networks,'' in \emph{Advances in Neural Information
  Processing Systems}, F.~Pereira, C.~J.~C. Burges, L.~Bottou, and K.~Q.
  Weinberger, Eds., vol.~25.\hskip 1em plus 0.5em minus 0.4em\relax Curran
  Associates, Inc., 2012. [Online]. Available:
  \url{https://proceedings.neurips.cc/paper/2012/file/c399862d3b9d6b76c8436e924a68c45b-Paper.pdf}
\BIBentrySTDinterwordspacing

\bibitem{SimonyanZ14a}
\BIBentryALTinterwordspacing
K.~Simonyan and A.~Zisserman, ``Very deep convolutional networks for
  large-scale image recognition,'' in \emph{3rd International Conference on
  Learning Representations, {ICLR} 2015, San Diego, CA, USA, May 7-9, 2015,
  Conference Track Proceedings}, Y.~Bengio and Y.~LeCun, Eds., 2015. [Online].
  Available: \url{http://arxiv.org/abs/1409.1556}
\BIBentrySTDinterwordspacing

\bibitem{7298594}
C.~Szegedy, W.~Liu, Y.~Jia, P.~Sermanet, S.~Reed, D.~Anguelov, D.~Erhan,
  V.~Vanhoucke, and A.~Rabinovich, ``Going deeper with convolutions,'' in
  \emph{2015 IEEE Conference on Computer Vision and Pattern Recognition
  (CVPR)}, 2015, pp. 1--9.

\bibitem{hochreiter1997long}
S.~Hochreiter and J.~Schmidhuber, ``Long short-term memory,'' \emph{Neural
  computation}, vol.~9, no.~8, pp. 1735--1780, 1997.

\bibitem{650093}
M.~Schuster and K.~Paliwal, ``Bidirectional recurrent neural networks,''
  \emph{IEEE Transactions on Signal Processing}, vol.~45, no.~11, pp.
  2673--2681, 1997.

\bibitem{cho2014learning}
K.~Cho, B.~van Merrienboer, C.~Gulcehre, D.~Bahdanau, F.~Bougares, H.~Schwenk,
  and Y.~Bengio, ``Learning phrase representations using rnn encoder-decoder
  for statistical machine translation,'' 2014.

\bibitem{maape}
S.~Kim and H.~Kim, ``A new metric of absolute percentage error for intermittent
  demand forecasts,'' \emph{International Journal of Forecasting}, vol.~32,
  no.~3, pp. 669--679, 2016.

\bibitem{YAO2019276}
\BIBentryALTinterwordspacing
J.~Yao and A.~Moawad, ``Vehicle energy consumption estimation using large scale
  simulations and machine learning methods,'' \emph{Transportation Research
  Part C: Emerging Technologies}, vol. 101, pp. 276--296, 2019. [Online].
  Available:
  \url{https://www.sciencedirect.com/science/article/pii/S0968090X19302293}
\BIBentrySTDinterwordspacing

\bibitem{de_cauwer_data-driven_2017}
\BIBentryALTinterwordspacing
C.~De~Cauwer, W.~Verbeke, T.~Coosemans, S.~Faid, and J.~Van~Mierlo,
  ``\BIBforeignlanguage{en}{A {Data}-{Driven} {Method} for {Energy}
  {Consumption} {Prediction} and {Energy}-{Efficient} {Routing} of {Electric}
  {Vehicles} in {Real}-{World} {Conditions}},''
  \emph{\BIBforeignlanguage{en}{Energies}}, vol.~10, no.~5, p. 608, May 2017.
  [Online]. Available: \url{http://www.mdpi.com/1996-1073/10/5/608}
\BIBentrySTDinterwordspacing

\bibitem{MODI2020454}
\BIBentryALTinterwordspacing
S.~Modi, J.~Bhattacharya, and P.~Basak, ``Estimation of energy consumption of
  electric vehicles using deep convolutional neural network to reduce
  driver’s range anxiety,'' \emph{ISA Transactions}, vol.~98, pp. 454--470,
  2020. [Online]. Available:
  \url{https://www.sciencedirect.com/science/article/pii/S001905781930401X}
\BIBentrySTDinterwordspacing

\bibitem{ZHANG2020115408}
\BIBentryALTinterwordspacing
J.~Zhang, Z.~Wang, P.~Liu, and Z.~Zhang, ``Energy consumption analysis and
  prediction of electric vehicles based on real-world driving data,''
  \emph{Applied Energy}, vol. 275, p. 115408, 2020. [Online]. Available:
  \url{https://www.sciencedirect.com/science/article/pii/S030626192030920X}
\BIBentrySTDinterwordspacing

\bibitem{GENIKOMSAKIS201798}
\BIBentryALTinterwordspacing
K.~N. Genikomsakis and G.~Mitrentsis, ``A computationally efficient simulation
  model for estimating energy consumption of electric vehicles in the context
  of route planning applications,'' \emph{Transportation Research Part D:
  Transport and Environment}, vol.~50, pp. 98--118, 2017. [Online]. Available:
  \url{https://www.sciencedirect.com/science/article/pii/S1361920915302881}
\BIBentrySTDinterwordspacing

\bibitem{LOPEZ2020124188}
\BIBentryALTinterwordspacing
F.~C. López and R.~Álvarez Fernández, ``Predictive model for energy
  consumption of battery electric vehicle with consideration of
  self-uncertainty route factors,'' \emph{Journal of Cleaner Production}, vol.
  276, p. 124188, 2020. [Online]. Available:
  \url{https://www.sciencedirect.com/science/article/pii/S0959652620342335}
\BIBentrySTDinterwordspacing

\bibitem{ALWREIKAT2021117096}
\BIBentryALTinterwordspacing
Y.~Al-Wreikat, C.~Serrano, and J.~R. Sodré, ``Driving behaviour and trip
  condition effects on the energy consumption of an electric vehicle under
  real-world driving,'' \emph{Applied Energy}, vol. 297, p. 117096, 2021.
  [Online]. Available:
  \url{https://www.sciencedirect.com/science/article/pii/S0306261921005444}
\BIBentrySTDinterwordspacing

\bibitem{fiori_effect_2019}
\BIBentryALTinterwordspacing
C.~Fiori, V.~Arcidiacono, G.~Fontaras, M.~Makridis, K.~Mattas, V.~Marzano,
  C.~Thiel, and B.~Ciuffo, ``\BIBforeignlanguage{en}{The effect of electrified
  mobility on the relationship between traffic conditions and energy
  consumption},'' \emph{\BIBforeignlanguage{en}{Transportation Research Part D:
  Transport and Environment}}, vol.~67, pp. 275--290, Feb. 2019. [Online].
  Available:
  \url{https://www.sciencedirect.com/science/article/pii/S136192091830662X}
\BIBentrySTDinterwordspacing

\bibitem{XIE2020115081}
\BIBentryALTinterwordspacing
Y.~Xie, Y.~Li, Z.~Zhao, H.~Dong, S.~Wang, J.~Liu, J.~Guan, and X.~Duan,
  ``Microsimulation of electric vehicle energy consumption and driving range,''
  \emph{Applied Energy}, vol. 267, p. 115081, 2020. [Online]. Available:
  \url{https://www.sciencedirect.com/science/article/pii/S0306261920305936}
\BIBentrySTDinterwordspacing

\bibitem{qi_data-driven_2016}
X.~Qi and Y.~Zhang, ``Data-{Driven} {Macroscopic} {Energy} {Consumption}
  {Estimation} for {Electric} {Vehicles} with {Different} {Information}
  {Availability},'' in \emph{2016 {International} {Conference} on
  {Computational} {Science} and {Computational} {Intelligence} ({CSCI})}, Dec.
  2016, pp. 1214--1219.

\bibitem{QI201836}
\BIBentryALTinterwordspacing
X.~Qi, G.~Wu, K.~Boriboonsomsin, and M.~J. Barth, ``Data-driven decomposition
  analysis and estimation of link-level electric vehicle energy consumption
  under real-world traffic conditions,'' \emph{Transportation Research Part D:
  Transport and Environment}, vol.~64, pp. 36--52, 2018, the contribution of
  electric vehicles to environmental challenges in transport. WCTRS conference
  in summer. [Online]. Available:
  \url{https://www.sciencedirect.com/science/article/pii/S1361920916307714}
\BIBentrySTDinterwordspacing

\bibitem{YI2018344}
\BIBentryALTinterwordspacing
Z.~Yi, J.~Smart, and M.~Shirk, ``Energy impact evaluation for eco-routing and
  charging of autonomous electric vehicle fleet: Ambient temperature
  consideration,'' \emph{Transportation Research Part C: Emerging
  Technologies}, vol.~89, pp. 344--363, 2018. [Online]. Available:
  \url{https://www.sciencedirect.com/science/article/pii/S0968090X18302365}
\BIBentrySTDinterwordspacing

\bibitem{BASSO2019141}
\BIBentryALTinterwordspacing
R.~Basso, B.~Kulcsár, B.~Egardt, P.~Lindroth, and I.~Sanchez-Diaz, ``Energy
  consumption estimation integrated into the electric vehicle routing
  problem,'' \emph{Transportation Research Part D: Transport and Environment},
  vol.~69, pp. 141--167, 2019. [Online]. Available:
  \url{https://www.sciencedirect.com/science/article/pii/S1361920918304760}
\BIBentrySTDinterwordspacing

\bibitem{BASSO202124}
\BIBentryALTinterwordspacing
R.~Basso, B.~Kulcsár, and I.~Sanchez-Diaz, ``Electric vehicle routing problem
  with machine learning for energy prediction,'' \emph{Transportation Research
  Part B: Methodological}, vol. 145, pp. 24--55, 2021. [Online]. Available:
  \url{https://www.sciencedirect.com/science/article/pii/S0191261520304549}
\BIBentrySTDinterwordspacing

\bibitem{LIU201774}
\BIBentryALTinterwordspacing
K.~Liu, T.~Yamamoto, and T.~Morikawa, ``Impact of road gradient on energy
  consumption of electric vehicles,'' \emph{Transportation Research Part D:
  Transport and Environment}, vol.~54, pp. 74--81, 2017. [Online]. Available:
  \url{https://www.sciencedirect.com/science/article/pii/S1361920917303887}
\BIBentrySTDinterwordspacing

\end{thebibliography}
\bibliographystyle{IEEEtran}

% \vspace{-10mm} 
\begin{IEEEbiographynophoto}
{Ayman Moawad}
is a research engineer in the Vehicle and Mobility Simulation group at Argonne National Laboratory. He received a Master's degree in Mechatronics, Robotics, and Computer Science from the Ecole des Mines, France and a Master's degree in Statistics from the University of Chicago, USA. His research interests include engineering applications of artificial intelligence for energy consumption and cost prediction of advanced vehicles, machine learning, large scale data analysis, and high performance computing.
\end{IEEEbiographynophoto}

% \vspace{-14mm} 
\begin{IEEEbiographynophoto}
{Krishna Murthy Gurumurthy}
is a Computational Transportation Engineer in the Transportation Systems \& Mobility Group at the Argonne National Laboratory. He received his doctorate for research focusing on travel demand modeling and forecasting, especially through the use of large-scale agent-based simulation tools. He is particularly interested in capturing the impacts of shared and automated vehicles on travel patterns and congestion and measuring the resulting effects of various policies.
\end{IEEEbiographynophoto}

% \vspace{-14mm} 
\begin{IEEEbiographynophoto}
{Ömer Verbas}
is the Technical Lead for Network Modeling and Simulation at the Transportation Systems \& Mobility Group at Argonne National Laboratory. His primary research areas are in transportation network modeling; multi-modal routing, assignment, and simulation; transit network design and scheduling; and charging and routing behavior of drivers with electric vehicles (EV). Throughout his career at Argonne, Dr. Verbas has been working on several transit and EV-related tasks under the SMART Mobility Consortium, an effort led by the Department of Energy (DOE) and multiple National Laboratories that aims to deliver new data, analysis, and modeling tools, and create new knowledge to support smarter mobility systems. He has won the “Pacesetter Award” at Argonne for his work on transit models, the “Impact Argonne Award” once for his work for the SMART Mobility Consortium and another time for his work related to COVID impact.
\end{IEEEbiographynophoto}

% \vspace{-14mm} 
\begin{IEEEbiographynophoto}
{Zhijian Li}
is a Mathematics Ph.D. student at University of California, Irvine. He received his Bachelor's degree in Applied Mathematics from University of California, Los Angeles. His current research interest lies in compression of neural networks, including quantization-aware training, sparsification, and channel pruning. He has also worked on spatiotemporal time-series forecasting through graph-based recurrent neural networks (RNNs).
\end{IEEEbiographynophoto}

% \vspace{-14mm} 
\begin{IEEEbiographynophoto}
{Ehsan Islam}
completed his M.Sc. in Interdisciplinary Engineering from Purdue University, USA, in 2019 and B.A.Sc. in Mechatronics Engineering from the University of Waterloo, Canada, in 2016. He applies mechatronics principles to innovate processes in advanced vehicle technologies and controls systems. At Argonne National Laboratory, he focuses his research on vehicle energy consumption analyses and inputs for DOE-VTO and NHTSA/U.S. Environmental Protection Agency/U.S. Department of Transportation CAFE and CO$_{2}$ standards using innovative large-scale simulation processes and applications of artificial intelligence.
\end{IEEEbiographynophoto}

% \vspace{-14mm} 
\begin{IEEEbiographynophoto}
{Vincent Freyermuth}
is a research engineer in the Transportation Systems Simulation and Vehicle and Mobility Simulation Group at Argonne National Laboratory. Vincent has 20 years of experience in the area of vehicle simulation, vehicle integration and product development. Vincent links agent-based transportation level models (POLARIS) to detailed vehicle level models (Autonomie) and life-cycle greenhouse gas analysis tool (GREET) to understand the broader impact of emerging mobility trends.
\end{IEEEbiographynophoto}

% \vspace{-14mm} 
\begin{IEEEbiographynophoto}
{Aymeric Rousseau}
is manager of the Vehicle and Mobility Simulation group at Argonne National Laboratory. He received his engineering diploma at the Industrial System Engineering School, France, in 1997 and an Executive MBA from Chicago Booth School of Business in 2019. For the past 20 years, he has been evaluating the impact of advanced vehicle and transportation technologies from a mobility and energy point of view, including the development of Autonomie (vehicle system simulation) and POLARIS (large-scale transportation system simulation).
\end{IEEEbiographynophoto}

\end{document}